\documentclass[conference]{IEEEtran}
\IEEEoverridecommandlockouts
\usepackage{cite}
\usepackage{amsmath,amssymb,amsfonts}
\usepackage{graphicx, caption, subcaption}
\usepackage{textcomp}
\usepackage{xcolor}
\usepackage{algorithm}
\usepackage{algpseudocode}
\usepackage{soul}
\usepackage{placeins}

\usepackage{amsthm}
\newtheorem{theorem}{Theorem}
\newtheorem{definition}{Definition}

\newtheorem{lemma}{Lemma}
\newtheorem{assumption}{Assumption}

\newtheorem{corollary}{Corollary}

\newcommand{\alg}{$\mathsf{DIAMOND}~$}
\newcommand{\algns}{$\mathsf{DIAMOND}$}

\def\BibTeX{{\rm B\kern-.05em{\sc i\kern-.025em b}\kern-.08em
    T\kern-.1667em\lower.7ex\hbox{E}\kern-.125emX}}

\renewcommand{\a}{\mathbf{a}}

\renewcommand{\b}{\mathbf{b}}

\renewcommand{\c}{\mathbf{c}}

\newcommand{\e}{\mathbf{e}}

\newcommand{\g}{\mathbf{g}}

\newcommand{\I}{\mathbf{I}}

\newcommand{\M}{\mathbf{M}}

\newcommand{\p}{\mathbf{p}}

\renewcommand{\u}{\mathbf{u}}
\newcommand{\V}{\mathbf{V}}

\renewcommand{\v}{\mathbf{v}}

\newcommand{\x}{\mathbf{x}}

\newcommand{\y}{\mathbf{y}}

\newcommand{\0}{\mathbf{0}}
\newcommand{\1}{\mathbf{1}}

\begin{document}


\title{DIAMOND: Taming Sample and Communication Complexities in Decentralized Bilevel Optimization}



\author{Peiwen Qiu$^{1}$ \mbox{\hspace{0.4cm}} Yining Li$^{1}$ \mbox{\hspace{0.4cm}} Zhuqing Liu$^{1}$ \mbox{\hspace{0.4cm}} Prashant Khanduri$^{2}$ \mbox{\hspace{0.4cm}} Jia Liu$^{1}$
\\ \mbox{\hspace{0.4cm}} Ness B. Shroff$^{1}$ \mbox{\hspace{0.4cm}} Elizabeth Serena Bentley$^{3}$ \mbox{\hspace{0.4cm}} Kurt Turck$^{3}$
\\ $^{1}$Dept. of ECE, The Ohio State University\mbox{\hspace{0.4cm}}
$^{2}$Dept. of CS, Wayne State University
\\ $^{3}$Air Force Research Laboratory, Information Directorate
\thanks{
This work has been supported in part by NSF grants CAREER CNS-2110259, CNS-2112471, CNS-2102233, CCF-2110252, and AFRL grant FA8750-20-3-1003. The views and conclusions contained herein are those of the authors and should not be interpreted as necessarily representing the official policies or endorsements, either expressed or implied, of the Air Force Research Laboratory or the U.S. Government.
Distribution A. Approved for public release: Distribution unlimited AFRL-2023-0273 on 18 Jan 2023.
}
}

\maketitle

\begin{abstract}
Decentralized bilevel optimization has received increasing attention recently due to its foundational role in many emerging multi-agent learning paradigms (e.g., multi-agent meta-learning and multi-agent reinforcement learning) over peer-to-peer edge networks.
However, to work with the limited computation and communication capabilities of edge networks, a major challenge in developing decentralized bilevel optimization techniques is to lower sample and communication complexities.
This motivates us to develop a new decentralized bilevel optimization called \alg (\ul{d}ecentralized s\ul{i}ngle-timescale stoch\ul{a}stic approxi\ul{m}ation with m\ul{o}me\ul{n}tum and gra\ul{d}ient-tracking).
The contributions of this paper are as follows:
i) our \alg algorithm adopts a single-loop structure rather than following the natural double-loop structure of bilevel optimization, which offers low computation and implementation complexity;
ii) compared to existing approaches, the \alg algorithm does not require any full gradient evaluations, which further reduces both sample and computational complexities;
iii) through a careful integration of momentum information and gradient tracking techniques, we show that the \alg algorithm enjoys $\mathcal{O}(\epsilon^{-3/2})$ in sample and communication complexities for achieving an $\epsilon$-stationary solution, both of which are independent of the dataset sizes and significantly outperform existing works.
Extensive experiments also verify our theoretical findings.
\end{abstract}




\section{Introduction} \label{sec: Introduction}


In recent years, the problem of performing decentralized bilevel optimization over networks has attracted increasing attention.
For a peer-to-peer communication network represented by a graph $\mathcal{G}=(\mathcal{N,L})$, where $\mathcal{N}$ and $\mathcal{L}$ denote the sets of agents and edges with $|\mathcal{N}|=m$, a decentralized bilevel optimization problem can be generally written as follows:
\begin{align}
\label{eq:dec_bilevel_prob}
        &\min _{\x \in \mathbb{R}^{d_{\mathrm{up}}}} l\left( \x \right)=\frac{1}{m} \sum_{i=1}^{m}\left\{ l_{i}\left(\x \right)\triangleq 
        \mathbb{E}_{\xi_{i}}\left[f_{i}\left(\x, \y_{i}^{*}\left(\x \right); \xi_{i}\right)\right] \right\}\text{,} \nonumber\\
        &\text{s.t. }\y_{i}^{*}\left(\x \right)=\arg\!\!\!\!\min_{\!\!\!\!\!\!\!\y_{i} \in \mathbb{R}^{d_{\mathrm{low}}}} \!\! \left\{ g_{i}\left(\x, \y_{i}\right)\triangleq \mathbb{E}_{\zeta_{i}}\left[g_{i}\left(\x_{i}, \y_{i}; \zeta_{i}\right)\right] \right\}, \!\!\!
\end{align}
where $l_{i}\left(\x \right)$ is the local objective function at agent $i$,
$\x \in \mathbb{R}^{d_{\mathrm{up}}}$ and $\y_{i} \in \mathbb{R}^{d_{\mathrm{low}}}$ are the global upper-level variables and the local lower-level variables at agent $i\in \left\{ 1,\ldots,m \right\}$,
and $\xi_{i}$ and $\zeta_{i}$ represent the random samples of the upper-level and the lower-level subproblems, respectively. 

Problem~\eqref{eq:dec_bilevel_prob} plays a foundational role for various fundamental multi-agent learning paradigms over decentralized or multi-hop wireless edge networks.
For instance, in the well-known actor-critic framework for cooperative multi-agent reinforcement learning (MARL, e.g., \cite{hong2020two, zhang2020bi}), the shared global policy improvement (the actor component) corresponds to the upper-level subproblem in Problem~\eqref{eq:dec_bilevel_prob}, which depends on the optimal solution of a policy evaluation subproblem ( the critic component), which corresponds to the lower-level subproblem in Problem~\eqref{eq:dec_bilevel_prob}.
Another example can be found in multi-agent meta-learning (also referred to as ``learning-to-learn'', see, e.g., \cite{liu2021boml,rajeswaran2019meta}), where the training of task-specific parameters at each agent can be represented by the lower-level subproblem in Problem~\eqref{eq:dec_bilevel_prob}.
The task-specific parameter training is coupled with the shared parameters' optimization, which corresponds to the upper-level subproblem in Problem~\eqref{eq:dec_bilevel_prob}.



It is evident from Problem~\eqref{eq:dec_bilevel_prob} that the most prominent features of decentralized bilevel optimization are i) ``bilevel'' and ii) ``decentralization.''
Same as the single-agent bilevel counterpart\cite{dempe2002foundations}, Problem~\eqref{eq:dec_bilevel_prob} has a hierarchical structure, where the upper-level subproblem objective value is determined by both the upper-level variable $\x$ and the optimal variables $\{ \y_i^*(\x) \}_{i=1}^{m}$ obtained by solving the lower-level subproblems.
Due to this bilevel structure, the solution approach for Problem~\eqref{eq:dec_bilevel_prob} is fundamentally different from the traditional loss minimization in conventional learning problems with a single-level structure.
Thus, new algorithm design and analysis techniques are necessary for solving Problem~\eqref{eq:dec_bilevel_prob}.
Moreover, instead of only having a single task as in single-agent bilevel optimization problems, one needs to cope with {\em multiple} lower-level tasks in a {\em decentralized} fashion in Problem~\eqref{eq:dec_bilevel_prob}.
This key difference necessitates
new algorithm designs. 
Thus, solving decentralized bilevel optimization problems over wireless networks needs to address the following technical challenges:

\begin{list}{\labelitemi}{\leftmargin=1em \itemindent=-0.5em \itemsep=.2em}

\item {\em Single-Loop or Double-Loop Architecture?} In Problem~\eqref{eq:dec_bilevel_prob}, it is often impractical to asymptotically solve the lower-level problem to optimality.
Rather, one typically resorts to using an {\em approximaion} of $\y^*(\x)$, which is obtained by solving the lower-level problems with finite iterations\cite{ghadimi2018approximation, ji2021bilevel, yang2021provably, chen2021single, guo2021randomized, hong2020two, khanduri2021near}.
However, due to the coordination complexity among agents and training accuracy concerns,
the algorithmic architecture choice between ``double-loop'' \cite{ghadimi2018approximation, ji2021bilevel, yang2021provably} or ``single-loop'' \cite{chen2021single, guo2021randomized, hong2020two, khanduri2021near}, both of which are widely used in single-agent bilevel optimization, suddenly becomes critical.
On one hand, the double-loop architecture naturally follows the bilevel problem structure and executes multiple inner-loop iterations within each outer iteration, which typically yields a more accurate estimation of the lower-level minimizer.
However, in a decentralized network setting, this requires a two-timescale implementation with high coordination complexity, as well as high computation 
and sample complexities in the inner loop.
On the other hand, the single-loop architecture iteratively solves the upper- and lower-level subproblems and updates corresponding parameters simultaneously, which is much easier to implement in the decentralized setting.
However, it is unclear whether the less accurate inner subproblem solutions could also result in high communication and sample complexities in the overall training process.

\smallskip
\item {\em Achieving Low Communication and Sample Complexities:} Since there is no dedicated centralized server in decentralized bilevel optimization over edge networks, it is infeasible to aggregate the local datasets at the geographically dispersed agents.
Hence, it is necessary for the agents to communicate and exchange information with each other to reach a ``consensus solution''~\cite{zhang2021low, Liu2022interact, nedic2009distributed, nedic2010constrained}.
In such scenarios, how to design efficient algorithms to reduce the required rounds of communications (i.e., communication complexity) to reach consensus is one of the most important questions in algorithm design.
This is particularly true for deploying decentralized bilevel optimization over wireless edge networks that may have low-speed and unreliable links.
Also, due to the fact that the agents (i.e., computing nodes) in many wireless edge networks are fundamentally constrained by computation capabilities (e.g., sensors or smart phones with limited computation speed, energy, and memory), it is important to design efficient algorithms to reduce the required number of samples (i.e., sample complexity).
However, achieving low sample and communication complexities are two fundamentally {\em conflicting}
goals. 
On one hand, the variance of a stochastic gradient highly depends on the number of samples in each mini-batch.
The more samples in each mini-batch (i.e., potentially higher sample complexity), the larger the variance, which may imply fewer communication rounds for convergence (i.e., lower communication complexity).
On the other hand, if one prefers to
use fewer samples per iteration to lower per-iteration sample complexity,
the stochastic gradient information is noisier, which could result in
more communication rounds to reach certain training accuracy
(i.e., higher communication complexity). 

\end{list}

In addition to the above challenges, the coupled structure and the inherent non-convexity of the decentralized bilevel optimization problems make it challenging to design algorithms and theoretically analyze their performance.
So far, results on designing decentralized bilevel optimization algorithms with low sample, communication, and implementation complexities remains rather limited in the literature (see Section~\ref{sec: RelatedWork} for more detailed discussions).
The main contribution of this paper is that we propose a series of new algorithm design techniques, which overcome the aforementioned challenges and achieve low sample and communication complexities with a {\em single-loop} structure for decentralized bilevel optimization problems.
The key results of this paper are summarized as follows:

\begin{list}{\labelitemi}{\leftmargin=1em \itemindent=-0.5em \itemsep=.2em}

\item We propose an algorithm called \alg (\ul{d}ecentralized s\ul{i}ngle-loop stoch\ul{a}stic approxi\ul{m}ation with m\ul{o}me\ul{n}tum and gra\ul{d}ient-tracking) for solving decentralized bilevel optimization problems over networks. 
Our proposed \alg algorithm integrates consensus-based updates with gradient tracking and momentum-based stochastic gradient estimators, which is a carefully designed triple-hybrid approach.
We show that this triple-hyrid approach enables the use of a single-loop algorithmic architecture, which significantly reduces the implementation complexity over peer-to-peer edge networks.
    
\item We show that \alg achieves a sample complexity of $\mathcal{O}(\epsilon^{-3/2})$ to find an $\epsilon$-stationary solution for non-convex upper-level optimization objectives.
Interestingly, this result matches the state-of-the-art sample complexity of stochastic first-order algorithms for solving single-agent bilevel optimization problems.
Meanwhile, the communication complexity of \alg is $\mathcal{O}(\epsilon^{-3/2})$.
These results show that \alg strikes a good balance between sample and communication complexities.
    
\item We conduct extensive experiments to validate the theoretical results of the proposed \alg algorithm. Our experimental results show that \alg outperforms other stochastic first-order baseline algorithms in terms of sample complexities in various communication network settings.
   
\end{list}

The rest of the paper is organized as follows. 
In Section~\ref{sec: RelatedWork}, we review related work to provide the necessary background on decentralized and bilevel optimization, and put our work in comparative perspectives.
In Section~\ref{sec: Preliminary}, we present the system model and the consensus reformulation of decentralized bilevel optimization. 
In Section~\ref{sec: Algorithm}, we propose the \alg algorithm.
We then provide the theoretical convergence analysis of \alg in Section~\ref{sec: ConvergenceAnalysis}. 
Section~\ref{sec: NumericalResults} provides numerical results to verify our theoretical findings, and Section~\ref{sec: Conclusion} concludes this paper.


\section{Related Work} \label{sec: RelatedWork}

To facilitate our discussions in subsequent sections, we organize the related work in three parts. 
First, we survey the approaches for solving single-agent bilevel optimization to provide a contrasting view for decentralized bilevel optimization. 
Then, we review the literature on decentralized optimization for single-level loss minimization to familiarize readers with the basics of decentralized optimization over networks. Lastly, we provide an in-depth comparison with the most related work on decentralized bilevel optimization.

\smallskip
\textbf{1)~Single-Agent Bilevel Optimization:}
{\em 1-a)~Gradient-Based Approaches:}
To our knowledge, single-agent bilevel optimization was first studied in~\cite{bracken1973mathematical}. 
Since then, several solution approaches have been proposed, such as 
1) penalizing the outer function with the optimality conditions of the inner problem~\cite{shi2005extended, mehra2021penalty}; 
2) reformulating the bilevel problem as a single-level problem by replacing the lower-level problem with its optimality conditions~\cite{colson2007overview, kunapuli2008classification};
and 3) utilizing gradient-based techniques to iteratively approximate the (stochastic) gradient of the upper-level problem.
Gradient-based algorithms for bilevel optimization have gained the most attention due to their ease of analysis.
Many gradient-based bilevel optimization algorithms have been proposed, including but not limited to: 
i) AID-based~\cite{ji2021lower, rajeswaran2019meta, shaban2019truncated},
ii) ITD-based~\cite{ghadimi2018approximation, gould2016differentiating, pedregosa2016hyperparameter},
and iii) SGD-based~\cite{ji2021bilevel, hong2020two, khanduri2021near, yang2021provably}.
However, these algorithms were designed for single-agent bilevel optimization problems and not applicable for the decentralized settings. 

{\em 1-b)~Momentum-Based Approaches:}
Momentum-based approaches enhance the gradient-based algorithms for single-agent bilevel optimization.
It has been shown that momentum improves the computation efficiency of stochastic gradient updates both in theory and in practice.
Several bilevel optimization algorithms that exploit momentum have been proposed, such as STABLE~\cite{chen2021single}, RSVRB~\cite{guo2021randomized}, MRBO~\cite{yang2021provably}, and SUSTAIN~\cite{khanduri2021near}. 
All of them share a similar single-loop algorithmic architecture. 
To reach an $\epsilon$-stationary point, STABLE requires an order of $\mathcal{O}\left( \epsilon^{-2} \right)$ samples, while RSVRB, MRBO and SUSTAIN require $\mathcal{O}\left( \epsilon^{-3/2} \right)$ samples.
Compared to STABLE, which only uses a momentum-based stochastic gradient estimator for the upper-level subproblems, RSVRB, MRBO, and SUSTAIN all utilize momentum-based stochastic gradient estimators for both the upper- and lower-level subproblems.
However, all these momentum-based algorithms are designed for the single-agent bilevel optimization setting. 
In comparison, we focus on decentralized multi-agent bilevel optimization, and propose the \alg algorithm, which uses momentum-based stochastic gradient estimators in both upper- and lower-levels.
Our theoretical result shows that \alg has the sample complexity of $\mathcal{O}\left( \epsilon^{-3/2} \right)$, matching the state-of-the-art result achieved by RSVRB, MRBO and SUSTAIN, but for the {\em more challenging decentralized setting.}


\smallskip
\textbf{2) Decentralized Optimization for Single-Level Loss Minimization over Networks:}
Decentralized optimization for single-level loss minimization over networks traces its roots to the seminal work\cite{tsitsiklis1984problems}, and has
found important applications in many engineering fields, e.g., network resource allocation~\cite{jiang2018consensus}, power networks~\cite{callaway2010achieving}, and robotic networks~\cite{kober2013reinforcement}.
One of the most popular methods to solve decentralized
optimization problems is
the distributed stochastic gradient descent (DSGD) \cite{nedic2009distributed}, which established the well-known $O(1/\sqrt{T})$ convergence rate with $T$ iterations.
Subsequently,
\cite{lian2017can} showed that DSGD can outperform the centralized SGD counterpart.
Recently, various sample- and communication-efficient variants of DSGD have been proposed, e.g., leveraging compression\cite{zhang2019compressed}, momentum~\cite{yu2019linear}, gradient tracking~\cite{lu2019gnsd, pu2021distributed}, and  variance reduction techniques~\cite{zhang2021low, xin2021hybrid,khanduri2021stem}.
However, results on solving general decentralized bilevel optimization problems are still limited.

\smallskip
\textbf{3) Decentralized Bilevel Optimization over Networks:}
So far, the research on decentralized bilevel optimization remains in its infancy.
As mentioned earlier, due to the lack of a centralized server in the decentralized setting, it is natural for us to consider
the network-consensus approach\cite{zhang2021low, Liu2022interact, nedic2009distributed, nedic2010constrained} as the solution strategy in this paper.
To our knowledge, the most related and the only work that also adopts a consensus-based approach for solving decentralized bilevel optimization problems is reported in \cite{Liu2022interact}, which contains two algorithmic variants called INTERACT and SVR-INTERACT.
Specifically, INTERACT is a local-full-gradient-based algorithm with gradient tracking and achieves $\left[ \mathcal{O}\left( n\epsilon^{-1} \right), \mathcal{O}\left( \epsilon^{-1} \right) \right]$ sample-communication complexity, where $n$ is the size of the dataset at each agent.
Recall that the sample and communication complexities of our \alg are both $\mathcal{O}\left( \epsilon^{-3/2} \right)$, which is {\em independent} of data size. 
This implies that in the large dataset regime $\Theta(\epsilon^{-1/2})$, which is not uncommon in the era of ``big data,'' INTERACT suffers a higher sample complexity than that of our \alg algorithm.

To lower the sample complexity of INTERACT, SVR-INTERACT leverages variance reduction techniques to retain the same communication complexity as INTERACT, while achieving a lower but still {\em dataset-dependent} sample complexity.
Thus, in the large dataset regime $\Omega(\epsilon^{-1})$, the sample complexity of SVR-INTERACT will be higher than that of \algns.
Also, SVR-INTERACT still requires periodic full gradients, while \alg only needs stochastic gradient evaluations.
Moreover, due to the variance reduction techniques, SVR-INTERACT has a double-loop algorithmic architecture.
By contrast, \alg is {\em single-loop} structure, which has a lower computational cost and is easier to implement.
Apart from these, SVR-INTERACT was designed to solve deterministic bilevel optimization problems, rather than stochastic bilevel optimization problems that has wider applications when the sample size of training data is large (e.g., hyperparameter optimization~\cite{franceschi2018bilevel}) or the fresh data is sampled for algorithm iterations (e.g., reinforcement learning~\cite{hong2020two}).





\section{System Model and Problem Formulation} \label{sec: Preliminary}
In this section, we will present the network-consensus-based problem reformulation that paves the way for our subsequent algorithm design and analysis. 
%
Recall from Problem~\eqref{eq:dec_bilevel_prob} that we consider a peer-to-peer communication network represented by a graph. 
Suppose that each agent $i$ can communicate with its set of neighbors denoted by $\mathcal{N}_{i}\triangleq \left\{ j\in \mathcal{N},:\left( i,j \right)\in \mathcal{L} \right\}$.
To solve Problem~\eqref{eq:dec_bilevel_prob} in a decentralized fashion, one can rewrite Problem~\eqref{eq:dec_bilevel_prob} as follows:
\begin{align}
\label{eq:dec_bilevel_prob_consensus}
        &\min _{\x_{i} \in \mathbb{R}^{d_{\mathrm{up}}}} l\left( \x \right)=\frac{1}{m} \sum_{i=1}^{m}\left\{ l_{i}\left(\x_{i}\right)\triangleq 
        \mathbb{E}_{\xi_{i}}\left[f_{i}\left(\x_{i}, \y_{i}^{*}\left(\x_{i}\right); \xi_{i}\right)\right] \right\}\text{,} \nonumber\\
        &\text{s.t. }\y_{i}^{*}\left(\x_{i}\right)=\arg\!\!\!\!\min_{\y_{i} \in \mathbb{R}^{d_{\mathrm{low}}}} \left\{ g_{i}\left(\x_{i}, \y_{i}\right)\triangleq \mathbb{E}_{\zeta_{i}}\left[g_{i}\left(\x_{i}, \y_{i}; \zeta_{i}\right)\right] \right\}\text{,} \nonumber\\
        &\qquad \x_{i}=\x_{j} \text{, if } (i, j) \in \mathcal{L},
\end{align}
where $\x_{i}\in \mathbb{R}^{d_{\mathrm{up}}}$, $\forall i$, are the local copies of the global upper-level variables at agent $i\in \left\{ 1,\ldots,m \right\}$ and $g_i(\x_i,\y_i)$ is the local lower-level loss at agent $i$. 
For notation simplicity, we denote $f_{i}\left(\x_{i}, \y_{i}^{*}\left(\x_{i}\right)\right)\!\triangleq\!\mathbb{E}_{\xi_{i}}\left[f_{i}\left(\x_{i}, \y_{i}^{*}\left(\x_{i}\right); \xi_{i}\right)\right]$.
The equality constraint in  Problem~\eqref{eq:dec_bilevel_prob_consensus} ensures that all agents share the same global $\x$-value to achieve the minimization of the upper-level function, hence the name ``consensus form''~\cite{nedic2009distributed, nedic2010constrained, zhang2021low, Liu2022interact}.
We assume that $f_i( \x_i, \y_i^*(\x_i))$ is non-convex in $\x_i$, $\forall i$, in general, and $g_i(\x_i,\y_i)$ is strongly-convex in $\y_i$, $\forall i$, which typically holds in  meta-learning, hyper-parameter optimization (see Section~\ref{sec: NumericalResults}), and MARL with quadratically-regularized linear critics.
Now, we define the notion of $\epsilon$-stationarity that serves as the convergence metric.
\begin{definition}[$\epsilon$-Stationary Point] \label{def:stationary_point}
    A stochastic algorithm reaches an $\epsilon$-stationary point $\left\{ \x_{i},\y_{i},\forall i\in\left[ m \right]  \right\}$ if 
    \begin{equation*}
        \mathbb{E}\Big[ \underbrace{\|\nabla l(\bar{\x})\|^{2}}_{\substack{\mathrm{Stationarity} \\ \mathrm{Error} }}+\underbrace{\left\|\y^{*}-\y\right\|^{2}}_{\substack{\mathrm{Lower-Level} \\ \mathrm{Error}}}+
        \underbrace{\frac{1}{m} \sum\nolimits_{i=1}^{m}\left\|\x_{i}-\bar{\x}\right\|^{2}}_{\substack{\mathrm{Consensus}~ \mathrm{Error}}} \Big] \leq \epsilon\text{,}
    \end{equation*}
    where $\bar{\x}\triangleq \frac{1}{m} \sum_{i=1}^{m} \x_{i}$, $\y\triangleq  \left[\y_{1}^{\top}, \ldots, \y_{m}^{\top}\right]^{\top}$, and $\y^{*} \triangleq  \left[\y_{1}^{* \top}, \ldots, \y_{m}^{* \top}\right]^{\top}$.
    The expectation is taken over the randomness of the algorithm. 
\end{definition}

Next, we formally define the sample complexity and communication complexity of a decentralized algorithm, which are also used in the literature~(e.g., \cite{Liu2022interact, zhang2021low}).
\begin{definition}[Sample Complexity] 
The sample complexity is defined as the total number of incremental first-order oracle (IFO) calls required per node for an algorithm to reach an $\epsilon$-stationary point defined in Definition~\ref{def:stationary_point}, where one IFO call is defined as the evaluation of the stochastic gradient of upper- and lower-level problems at agent $i\in\left[ m \right]$.
\end{definition}

\begin{definition}[Communication Complexity] 
The communication complexity is defined as the total rounds of communications required to find an $\epsilon$-stationary point defined in Definition~\ref{def:stationary_point}, where each node can send and receive local parameters with its neighbors in one communication round.
\end{definition}





\section{The \alg Algorithm} \label{sec: Algorithm}
In this section, we present our \alg algorithm for solving Problem~\eqref{eq:dec_bilevel_prob_consensus}. 
Since the agents can communicate with their neighbors through the network to send and receive information (model parameters) and aggregate the received information, we define the consensus weight matrix $\mathrm{\M}\in \mathbb{R}^{m\times m}$, where $\left[\mathrm{\M}\right]_{ij}$ represents the consensus weight over edge $(i,j)\in\mathcal{L}$.
We assume that $\M$ satisfies the following:
\begin{list}{\labelitemi}{\leftmargin=1em \itemindent=-0.5em \itemsep=.2em}
    \item[1)] Doubly stochastic: $\sum_{i=1}^{m}\left[ \mathrm{\M} \right]_{ij}=\sum_{j=1}^{m}\left[ \mathrm{\M} \right]_{ij}=1, \forall i,j$;
    \item[2)] Symmetric: $\left[ \mathrm{\M} \right]_{ij}=\left[ \mathrm{\M} \right]_{ji}, \forall i,j\in \mathcal{N}$;
    \item[3)] Network-defined sparsity: $\left[ \mathrm{\M} \right]_{ij}>0$ if $\left( i,j \right)\!\in\! \mathcal{L}$; otherwise, $\left[ \mathrm{\M} \right]_{ij}=0, \forall i,j\in \mathcal{N}$.
\end{list}
The above conditions imply that the eigenvalues of $\mathrm{\M}$ are real and thus could be sorted as: $-1<\lambda_{m}\left( \mathrm{\M} \right)\le \cdots \le \lambda_{2}\left( \mathrm{\M} \right)<\lambda_{1}\left( \mathrm{\M} \right)=1$. 
We denote the second largest eigenvalue in magnitude of $\M$ as $\lambda\triangleq \max\left\{ \left| \lambda_{2}\left( \mathrm{\M} \right) \right|,\left| \lambda_{m}\left( \mathrm{\M} \right) \right| \right\}$, which will play an important role in the step-size selection for our proposed algorithm.
Note that the choice of $\M$ is not unique.
For example, one possible choice of $\mathrm{\M}$ that only relies on local information is the Metropolis weights~\cite{zhang2018fully}:
\begin{align*}
&[\mathrm{\M}]_{ij}=\left\{1+\max \left[d(i), d(j)\right]\right\}^{-1}, \forall(i, j) \in \mathcal{L}, \\
&[\mathrm{\M}]_{ii}=1-\sum_{j\neq i}[\mathrm{\M}]_{ij}, \forall i \in \mathcal{N},
\end{align*}
where $d(i)=|\mathcal{N}_i|$ is the degree of agent $i$. 

Using the implicit function theorem, the hypergradient of $l_{i}\left( \x_{i} \right)$ for a given $\x_{i}\in \mathbb{R}^{d_{\mathrm{up}}}$ can be evaluated as~\cite{ghadimi2018approximation}:
\begin{align}\label{eq:hypergradient}
    \nabla l_{i}(\x_{i}) = &\nabla_{\x} f_{i}\left(\x_{i}, \y_{i}^{*}(\x_{i})\right)-\nabla_{\x \y}^{2} g_{i}\left(\x_{i}, \y_{i}^{*}(\x_{i})\right) \times\nonumber\\
    &\quad \left[\nabla_{\y \y}^{2} g_{i}\left(\x_{i}, \y_{i}^{*}(\x_{i})\right)\right]^{-1} \nabla_{\y} f_{i}\left(\x_{i}, \y_{i}^{*}(\x_{i})\right)\text{.}
\end{align}
Since obtaining $\y_{i}^{*}(\x_{i})$ in closed-form is difficult, $\bar{\nabla} f_{i}\left( \x_{i},\y_{i} \right)$ is used as a surrogate of $\nabla l_{i}\left( \x_{i} \right)$ at any $\left( \x_{i},\y_{i} \right)\in \mathbb{R}^{d_{\mathrm{up}}\times d_{\mathrm{low}}}$, which is defined as follows~\cite{ghadimi2018approximation}:
\begin{align}\label{eq:approx_hypergradient}
    \bar{\nabla} f_{i}\left( \x_{i},\y_{i} \right)=&\nabla_{\x} f_{i}\left( \x_{i},\y_{i} \right)-\nabla_{\x \y}^{2} g_{i}\left( \x_{i},\y_{i} \right) \times\nonumber \\
    &\left[ \nabla_{\y \y}^{2} g_{i}\left( \x_{i},\y_{i} \right) \right]^{-1}\nabla_{\y} f_{i}\left( \x_{i},\y_{i} \right)\text{.}
\end{align}
Note that the computation of $\bar{\nabla} f_{i}\left( \x_{i},\y_{i} \right)$ involves exact Hessian matrix inverse and gradient, which is cumbersome. 
To avoid this expensive computation, we adopt the biased stochastic gradient estimation of $\bar{\nabla} f_{i}\left( \x_{i},\y_{i} \right)$ defined as~\cite{khanduri2021near}:
\begin{align}\label{eq:sg_f}
        &\hat{\nabla} f_{i}\left( \x_{i},\y_{i};\bar{\xi}_{i} \right)=
        \nabla_{\x} f_{i}\left( \x_{i},\y_{i};\xi_{i} \right)-\frac{K}{L_{g}}\nabla_{\x \y}^{2} g_{i}\left( \x_{i},\y_{i};\zeta_{i}^{0} \right) \nonumber\\
        &\quad \times \prod_{j=1}^{k\left( K \right)}\left( I-\frac{\nabla_{\y \y}^{2} g_{i}\left( \x_{i},\y_{i};\zeta_{i}^{j} \right)}{L_{g}} \right)\nabla_{\y} f_{i}\left( \x_{i},\y_{i};\xi_{i} \right), \!\!\!\!\!
\end{align}
where $k\left( K \right)\sim \mathcal{U}\{ 0, \cdots, K-1\}$ is a uniform random variable chosen from $\left\{ 0,\cdots, K-1 \right\}$.
A total of $K+2$ independent samples are collected from the upper- and lower-level  distributions $\pi_f$ and $\pi_g$, respectively.
We denote all random variables needed in \eqref{eq:sg_f} as a $(K\!+\!3)$-tuple: $\bar{\xi}_{i}\triangleq \left\{ \xi_{i},\zeta_{i}^{0},\cdots ,\zeta_{i}^{K},k\left( K \right) \right\}$, where $\xi_{i}\sim \pi_{f}$, $\zeta_{i}^{j}\sim \pi_{g}\text{, } j=0,\cdots, K$.

The overall framework of the proposed \alg algorithm for solving the decentralized bilevel optimization problem in~\eqref{eq:dec_bilevel_prob_consensus} is summarized in Algorithm~\ref{alg}. 
\begin{algorithm}[ht!]
\caption{The \alg Algorithm.}\label{alg}
\begin{algorithmic}[1]
\State \textbf{Input:} Step-sizes $\alpha_{t}$, $\beta_{t}$.
Momentum coefficients $\eta_{t}$, $\gamma_{t}$.
\State \textbf{Initialization:} Let $\left\{ \x_{i,-1}, \y_{i,-1} \right\}_{i=1}^{m}=\left\{ \x_{-1},\y_{-1} \right\}$. Let $\left\{ \p_{i,-1} \right\}_{i=1}^{m}=\0$, and $\left\{ \v_{i,-1} \right\}_{i=1}^{m}=\0$
\For {$t=0$ to $T-1$}
\For {each agent $i\in \left[ m \right]$}
    \State Estimate the stochastic gradients $\p_{i,t}$ and $\v_{i,t}$ \\ \quad \quad \quad \quad using \eqref{eq:p_it} and \eqref{eq:v_it};
    \State Track the global gradient $\u_{i,t}$ using~\eqref{eq:u_it};
    \State Update the local model parameters $\x_{i,t+1}$ and \\ \quad \quad \quad \quad $\y_{i,t+1}$ using~\eqref{eq:x_update} and~\eqref{eq:y_update};
\EndFor
\EndFor
\end{algorithmic}
\end{algorithm}
Note that \alg adopts a {\em single-loop} structure, which reduces the computation and implementation complexities compared with the double-loop structure.
Meanwhile, \alg relies on consensus updates along with gradient tracking and uses momentum-based stochastic gradient estimators, so that it matches state-of-the-art convergence guarantees.
Thus, \alg consists of three parts: i) local stochastic gradient estimation, ii) global gradient tracking, and iii) consensus update with stochastic gradient descent (SGD). 
The details of each part are described as follows:

\smallskip
{\bf 1) Local Stochastic Gradient Estimation:}
Each agent estimates the momentum-based upper- and lower-level update directions $\p_{i,t}$ and $\v_{i,t}$ of the upper-level and lower-level functions, respectively, with its local stochastic gradients:
\begin{align}\label{eq:p_it}
    \p_{i,t}=& \hat{\nabla} f_{i}\left(\x_{i, t}, \y_{i, t};\bar{\xi}_{i}\right) \nonumber\\
    &+\left( 1-\eta_{t} \right)\left( \p_{i,t-1}- \hat{\nabla} f_{i}\left(\x_{i, t-1}, \y_{i, t-1};\bar{\xi}_{i}\right)\right), \\
    \label{eq:v_it} \v_{i, t}=&\nabla_{\y} g_{i}\left(\x_{i, t}, \y_{i, t};\zeta_{i}\right) \nonumber\\
    &+\left( 1-\gamma_{t} \right) \left( \v_{i, t-1}-\nabla_{\y} \g_{i}\left(\x_{i, t-1}, \y_{i, t-1};\zeta_{i}\right) \right),
\end{align}
where $\eta_{t}\in \left[ 0,1 \right]$ and $\gamma_{t}\in \left[ 0,1 \right]$ are the momentum coefficients.
Note that the avoidance of the full gradient computation implies a noisy gradient estimation. 
\alg utilizes momentum-based stochastic gradient estimations to improve the accuracy of the current gradient estimation,
which is similar to~\cite{cutkosky2019momentum, tran2019hybrid} for single-level stochastic optimization and \cite{khanduri2021near, yang2021provably} for single-agent bilevel optimization.

\smallskip
{\bf 2) Global Gradient Tracking:}
Each agent updates the global gradient $\u_{i,t}$ by averaging all its neighbors' gradient estimates $\u_{j,t-1}$, $j\in\mathcal{N}_i$, which is defined as:
\begin{equation}\label{eq:u_it}
    \u_{i, t}=\sum_{j \in \mathcal{N}_{i}}[\M]_{i j} \u_{j, t-1}+\p_{i,t}-\p_{i,t-1}.  
\end{equation}
We do not perform gradient tracking for  $\v_{i,t}$ since the lower-level $\y$-variables do not require consensus (cf. Problem~\eqref{eq:dec_bilevel_prob_consensus}).

\smallskip
{\bf 3) Consensus Update with Decentralized SGD:}
Each agent $i$ updates the upper-level parameters by computing a weighted average of its neighbors' local copies $\x_{j,t}$, $j\in\mathcal{N}_i$ and using the tracked global gradient $\u_{i,t}$ computed in 2):
\begin{equation}\label{eq:x_update}
    \x_{i,t+1}=\sum_{j \in \mathcal{N}_{i}}[\M]_{i j}\x_{j,t}-\alpha_{t}\u_{i,t},
\end{equation}
where $\alpha_{t}$ is the upper-level step-size.
The lower-level model parameters $\y_{i,t+1}$ are updated locally by using $\v_{i,t}$:
\begin{equation}\label{eq:y_update}
    \y_{i,t+1}=\y_{i,t}-\beta_{t}\v_{i,t},
\end{equation}
where $\beta_{t}$ is the lower-level step-size.
Again, note that consensus is only required for the upper-level $\x_i$-parameters.



\section{Theoretical Performance Analysis} \label{sec: ConvergenceAnalysis}
In this section, we establish the theoretical convergence guarantees for the \alg algorithm for solving the decentralized bilevel optimization problem in~\eqref{eq:dec_bilevel_prob_consensus}.
Before describing the proof details, we first highlight three major challenges in the convergence analysis in our \alg algorithm:

\smallskip
{\bf a) Momentum-Based Stochastic Gradient Estimation Error of the Lower-Level Subproblem:} 
Although the stochastic gradient estimation $\nabla_{\y}g_i(\x_{i,t},\y_{i,t},\zeta_i)$ of the lower-level objective function is unbiased, there exists a bias 
between $\v_{i,t}$ and the full gradient due to the randomness and the added momentum, which can be written as $\e_{i,t}^{g}\triangleq \v_{i,t}-\nabla_{\y} g_{i}\left(\x_{i,t}, \y_{i,t}\right)$.

\smallskip
{\bf b) Momentum-Based Stochastic Gradient Estimation Error of the Upper-Level Subproblem:}
The momentum-based stochastic gradient estimation error of the upper-level subproblem is caused by:
i) the randomness of the gradient estimation $\p_{i,t}$, 
ii) the added momentum,
and iii) the approximation $\y_{i,t}\approx \y_{i,t}^{*}\left( \x_{i,t} \right)$. 
We denote this error as: $    \e_{i,t}^{f}\triangleq \p_{i,t}-\bar{\nabla} f_{i}\left(\x_{i,t}, \y_{i,t}\right)-\b_{i,t}$.
where 
$
\b_{i,t}\triangleq \mathbb{E}_{\bar{\xi}_{i}}[ \hat{\nabla} f_{i}\left( \x_{i,t},\y_{i,t};\bar{\xi}_{i} \right)]-\bar{\nabla} f_{i}\left( \x_{i,t},\y_{i,t} \right)$ is the bias.

\smallskip
{\bf c) Consensus Error:}
\alg utilizes a decentralized consensus update for the upper-level model parameters as shown in \eqref{eq:u_it}, which inevitably introduces consensus errors.


\subsection{Main Convergence Results}
To quantify the convergence rate performance of \algns, we first define a new convergence metric specifically for the decentralized bilevel problem in \eqref{eq:dec_bilevel_prob_consensus}:
    \begin{equation}\label{eq:convergence_metric}
        \mathfrak{M}_{t} \triangleq \left\|\nabla l\left(\bar{\x}_{t}\right)\right\|^{2}+\left\|\x_{t}-1 \otimes \bar{\x}_{t}\right\|^{2}+\left\|\y_{t}^{*}-\y_{t}\right\|^{2}\text{,}
    \end{equation}
    where $\bar{\x}_{t}\triangleq \frac{1}{m} \sum_{i=1}^{m} \x_{i,t}$, $\x_{t}\triangleq  \left[\x_{1,t}^{\top}, \ldots, \x_{m,t}^{\top}\right]^{\top}$, $\y_{t}\triangleq  \left[\y_{1,t}^{\top}, \ldots, \y_{m,t}^{\top}\right]^{\top}$, and $\y_{t}^{*} \triangleq  \left[\y_{1,t}^{* \top}, \ldots, \y_{m,t}^{* \top}\right]^{\top}$.
The first term in~\eqref{eq:convergence_metric} measures the convergence of the agent-average $\bar{\x}_{t}$ to a stationary point.
The second term in~\eqref{eq:convergence_metric} quantifies the consensus error among local copies of the upper-level $\x_t$-parameters.
The third term in~\eqref{eq:convergence_metric} measures the approximation error caused by the finite iterations of the lower-level optimization across all agents.
Clearly, if $\mathfrak{M}_{t}\to 0$, we can conclude that the algorithm achieves three goals simultaneously: 1) achieving a stationary solution of the decentralized bilevel optimization problem in~\eqref{eq:dec_bilevel_prob_consensus}, 2) reaching consensus of upper-level model parameters across all agents, and 3) obtaining optimal solutions to the lower-level subproblem.

Next, we state the following assumptions that are useful for our convergence performance analysis:
\begin{assumption}[Upper-Level Objective]\label{asmp_up}
    $f_{i}\left( \x_{i},\y_{i} \right)$ satisfies:
\begin{list}{\labelitemi}{\leftmargin=1.3em \itemindent=-0.8em \itemsep=.2em}
        \item[{\em 1)}] For any $i\in \left\{ 1,\ldots,m \right\}$ and $\left( \x_{i},\y_{i} \right)\in \mathbb{R}^{d_{\mathrm{up}}}\times \mathbb{R}^{d_{\mathrm{low}}}$, $\nabla _{\x}f_{i}\left( \x_{i},\y_{i} \right)$ and $\nabla _{\y}f_{i}\left( \x_{i},\y_{i} \right)$ are Lipschitz continuous with constants $L_{f_{x}}\ge 0$ and $L_{f_{y}}\ge 0$, respectively.
        
        \item[{\em 2)}] For any $i\in \left\{ 1,\ldots,m \right\}$ and $\left( \x_{i},\y_{i} \right)\in \mathbb{R}^{d_{\mathrm{up}}}\times \mathbb{R}^{d_{\mathrm{low}}}$, we have $\left\| \nabla _{\y}f_{i}\left( \x_{i},\y_{i} \right) \right\|\le C_{f_{y}}$ for some constants $C_{f_{y}}\ge 0$.
\end{list}
\end{assumption}

\begin{assumption}[Lower-Level Objective]\label{asmp_lo}
    $g_{i}\left( \x_{i},\y_{i} \right)$ satisfies:
\begin{list}{\labelitemi}{\leftmargin=1.3em \itemindent=-0.8em \itemsep=.2em}
        \item[{\em 1)}] For any $\x_{i}\in \mathbb{R}^{d_{\mathrm{up}}}$ and $\y_{i}\in \mathbb{R}^{d_{\mathrm{low}}}$, $g_{i}\left( \x_{i},\y_{i} \right)$ is twice continuously differentiable with respect to $\left( \x_{i},\y_{i} \right)$.
        
        \item[{\em 2)}] For any $i\in \left\{ 1,\ldots,m \right\}$ and $\left( \x_{i},\y_{i} \right)\in \mathbb{R}^{d_{\mathrm{up}}}\times \mathbb{R}^{d_{\mathrm{low}}}$, $\nabla _{\y}g_{i}\left( \x_{i},\y_{i} \right)$ is Lipschitz continuous with constant $L_{g}\ge 0$.
        
        \item[{\em 3)}] For any $i\in \left\{ 1,\ldots,m \right\}$ and $\x_{i} \in \mathbb{R}^{d_{\mathrm{up}}}$, $g_{i}\left( \x_{i},\cdot  \right)$ is $\mu_{g}$-strongly convex with respect to $\y_{i}$ for some $\mu_{g}>0$.
        
        \item[{\em 4)}] For any $i\in \left\{ 1,\ldots,m \right\}$ and $\left( \x_{i},\y_{i} \right)\in \mathbb{R}^{d_{\mathrm{up}}}\times \mathbb{R}^{d_{\mathrm{low}}}$, $\nabla _{\x\y}^{2}g_{i}\left( \x_{i},\y_{i} \right)$ and $\nabla _{\y\y}^{2}g_{i}\left( \x_{i},\y_{i} \right)$ are Lipschitz continuous with constants $L_{g_{xy}}\ge 0$ and $L_{g_{yy}}\ge 0$, respectively.
        
        \item[{\em 5)}] For any $i\in \left\{ 1,\ldots,m \right\}$ and $\left( \x_{i},\y_{i} \right)\in \mathbb{R}^{d_{\mathrm{up}}}\times \mathbb{R}^{d_{\mathrm{low}}}$, we have $\left\| \nabla _{\x\y}^{2}g_{i}\left( \x_{i},\y_{i} \right) \right\|^{2}\le C_{g_{xy}}$ for some $C_{g_{xy}}> 0$.
\end{list}
\end{assumption}

\begin{assumption}[Stochastic Objectives]\label{asmp_sf}
Assumptions~\ref{asmp_up} and \ref{asmp_lo} hold for $f_{i}\left( \x_{i},\y_{i};\xi_{i} \right)$ and $g_{i}\left( \x_{i},\y_{i};\zeta_{i} \right)$, for all $\xi_{i}\in \mathrm{supp}\left( \pi_{f} \right)$ and $\zeta_{i}\in \mathrm{supp}\left( \pi_{g} \right)$ where $\mathrm{supp}\left( \pi \right)$ denotes the support of distribution $\pi$.
\end{assumption}

\begin{assumption}[Stochastic Gradients]\label{asmp_sg}
    For any $i \!\in\! \left\{ 1,\ldots,m \right\}$ and $\left( \x_{i},\y_{i} \right) 
    \!\in\! \mathbb{R}^{d_{\mathrm{up}}} \!\times\! \mathbb{R}^{d_{\mathrm{low}}}$, the gradient estimators $\hat{\nabla} f_{i}\left( \x_{i},\y_{i};\bar{\xi}_{i} \right)$ and $\nabla_{\y} g_{i}\left(\x_{i}, \y_{i};\zeta_{i}\right)$ satisfy:
\begin{list}{\labelitemi}{\leftmargin=1.3em \itemindent=-0.8em \itemsep=.2em}
        \item[{\em 1)}] There exists a constant $\sigma_{f} \geq 0$  such that $\mathbb{E}_{\bar{\xi}_{i}} [\|\hat{\nabla} f_{i}(\x_{i}, \y_{i}; \bar{\xi}_{i}) -\bar{\nabla} f_{i}(\x_{i}, \y_{i})-\b_{i}(\x_{i}, \y_{i})\|^{2} ] \leq \sigma_{f}^{2}$,
        where $\b_{i} ( \x_{i},\y_{i} ) \triangleq \mathbb{E}_{\bar{\xi}_{i}} [ \hat{\nabla} f_{i} ( \x_{i},\y_{i};\bar{\xi}_{i} )]-\bar{\nabla} f_{i} ( \x_{i},\y_{i} )$ is the bias in estimating $\bar{\nabla} f_{i} ( \x_{i},\y_{i} )$.
        
        \item[{\em 2)}] There exists a constant $\sigma_{g} \geq 0$ such that $\mathbb{E}_{\zeta_{i}}[\|\nabla_{\y} g_{i}(\x_{i}, \y_{i} ; \zeta_{i})-\nabla_{\y}  g_{i}(\x_{i}, \y_{i})\|^{2}] \leq \sigma_{g}^{2}$.
\end{list}
\end{assumption}
We note that all these assumptions are standard in the bilevel optimization literature (see, e.g., ~\cite{Liu2022interact, khanduri2021near, ghadimi2018approximation}).
Next, we state two lemmas on characterizing the Lipschitz constants of the hypergradient $\nabla l_i(\x_i)$ in~\eqref{eq:hypergradient}, the approximate gradient $\bar{\nabla}f_i(\x_i,\y_i)$ in~\eqref{eq:approx_hypergradient}, the optimal solution $\y_{i}^{*}$ of the lower-level problem, and the stochastic gradient estimator $\hat{\nabla} f_{i}\left( \x_{i},\y_{i};\bar{\xi}_{i} \right)$ in~\eqref{eq:sg_f}. These lemmas will be useful in our main convergence results.
\begin{lemma}[Ref.~\cite{ghadimi2018approximation}]\label{lemma_Lip}
     Under Assumptions~\ref{asmp_up} and \ref{asmp_lo}, we have
    \begin{align*}
            & \|\bar{\nabla} f_{i}(\x,\y)-\nabla l_{i}(\x)\| \leq L_{f} \left\|\y^{*}(\x)-\y\right\|, \\
            & \left\|\y_{i}^{*}\left(\x_{1}\right)-\y_{i}^{*}\left(\x_{2}\right)\right\| \leq L_{y}\left\|\x_{1}-\x_{2}\right\|, \\
            &\left\|\nabla l_{i}\left(\x_{1}\right)-\nabla l_{i}\left(\x_{2}\right)\right\| \leq L_{l}\left\|\x_{1}-\x_{2}\right\|,
    \end{align*}
    for all $i\in \left\{ 1,\ldots,m \right\}$, $\x,\x_{1},\x_{2}\in \mathbb{R}^{d_{\mathrm{up}}}$ and $\y\in \mathbb{R}^{d_{\mathrm{low}}}$, where the Lipschitz constants above are defined as:
    \begin{align*}
    & L_{f}=L_{f_{x}}+\frac{L_{f_{y}} C_{g_{x y}}}{\mu_{g}}+C_{f_{y}}\left(\frac{L_{g_{x y}}}{\mu_{g}}+\frac{L_{g_{y y}} C_{g_{x y}}}{\mu_{g}^{2}}\right), \\
    & L_{l}=L_{f}+\frac{L_{f} C_{g_{x y}}}{\mu_{g}}, \quad \text{ and } \quad
    L_{y}=\frac{C_{g_{xy}}}{\mu_{g}}.
    \end{align*}
\end{lemma}

\begin{lemma}[Ref.~\cite{khanduri2021near}]\label{lemma_Lip_sg}
     Under Assumptions~\ref{asmp_up}, \ref{asmp_lo} and \ref{asmp_sf}, for any $\left( \x_{1},\y_{1} \right) \text{, }\left( \x_{2},\y_{2} \right)\in \mathbb{R}^{d_{\mathrm{up}}}\times \mathbb{R}^{d_{\mathrm{low}}}$, we have 
    \begin{multline*}
        \mathbb{E}_{\bar{\xi}}\left\|\hat{\nabla} f_{i}\left(\x_{1}, \y_{1} ; \bar{\xi}\right)- \hat{\nabla} f_{i}\left(\x_{2}, \y_{2} ; \bar{\xi}\right)\right\| \\
        \leq L_{K}(\left\|\x_{1}-\x_{2}\right\|+\left\|\y_{1}-\y_{2}\right\|),
    \end{multline*}
    where $L_{K}=$
    \begin{equation*}
        2 L_{f_{x}}^{2}+\frac{6 C_{g_{x y}}^{2} L_{f_{y}}^{2} K}{2 \mu_{g} L_{g}-\mu_{g}^{2}}+\frac{6 C_{f_{y}}^{2} L_{g_{x y}}^{2} K}{2 \mu_{g} L_{g}-\mu_{g}^{2}}+\frac{6 C_{g_{x y}}^{2} C_{f_{y}}^{2} L_{g_{y y}}^{2} K^{3}}{\left(L_{g}-\mu_{g}\right)^{2}\left(2 \mu_{g} L_{g}-\mu_{g}^{2}\right)}\text{,}
    \end{equation*}
    and $K$ is the number of samples required to construct the stochastic gradient estimate in~\eqref{eq:sg_f}.
\end{lemma}
The following lemma says that the bias of the stochastic gradient estimator for the upper-level objective defined in~\eqref{eq:sg_f} decays exponentially fast with respect to the number of samples $K$ that is chosen to approximate the Hessian inverse.
\begin{lemma}[Ref.~\cite{hong2020two}]\label{lemma_Bias}
     Under Assumptions~\ref{asmp_up}--\ref{asmp_sf}, for any $i\in \left\{1,\ldots,m \right\}$ and $\left( \x,\y \right)\in \mathbb{R}^{d_{\mathrm{up}}}\times \mathbb{R}^{d_{\mathrm{low}}}$, the bias of the stochastic gradient estimator in~\eqref{eq:sg_f} satisfies:
    \begin{equation*}
        \|\bar{\nabla} f_{i}(\x_{i}, \y_{i})-\mathbb{E}[\hat{\nabla} f_{i}(\x_{i}, \y_{i} ; \bar{\xi}_{i})]\| \leq \frac{C_{g_{x y}} C_{f_{y}}}{\mu_{g}}\left(1-\frac{\mu_{g}}{L_{g}}\right)^{K},
    \end{equation*}
    where $K$ is the number of samples required to construct the stochastic gradient estimate in~\eqref{eq:sg_f}.
\end{lemma}

Now, based on the convergence metric defined in~\eqref{eq:convergence_metric}, we state the main convergence result of \alg in Theorem~\ref{thm1}:
\begin{theorem}[Convergence Rate of \algns]\label{thm1}
    Under Assumptions~\ref{asmp_up}--\ref{asmp_sg}, choose $K=\left(L_{g} / \mu_{g}\right) \log \left(C_{g_{x y}} C_{f_{y}} T / \mu_{g}\right)$. 
    Define $\alpha_{t}\triangleq \left( \omega+t \right)^{-1/3}$ for $ \omega\ge 2$, $\beta_{t}\triangleq c_{\beta} \alpha_{t}$, $\eta_{t+1}\triangleq c_{\eta} \alpha_{t}^{2}$, and $\gamma_{t+1}\triangleq c_{\gamma} \alpha_{t}^{2}$, where 
    \begin{align*}
    & c_{\beta}=\frac{8 L_{f}}{m L_{\mu_{g}}\bar{c}_{y}}, \quad
    c_{\eta}=\frac{6 L_{f} \bar{c}_{\eta}+m}{3 L_{f} m\left(1-3 \bar{c}_{u} \bar{c}_{\eta}\right)}, \nonumber\\
    & c_{\gamma}=\frac{1}{3 L_{f}}+8 L_{g}^{2} c_{\beta}^{2}+\bar{c}_{r}\left(\frac{2 c_{\beta} \bar{c}_{y}}{L_{\mu_{ g}}}+\frac{8 L_{K}^{2} c_{\beta}^{2}}{\bar{c}_{\eta}}+12 L_{K}^{2} \bar{c}_{u} c_{\beta}^{2}\right),
    \end{align*}
    and where
    \begin{align*}
    & \bar{c}_{y}\!=\!\min \bigg\{\sqrt{\frac{L_{f}}{2L_{y}^{2} m}}, \sqrt{\frac{L_{f}}{10 L_{y}^{2} m^{2}}}\bigg\}, \,\,\bar{c}_{u}\!=\!\min \bigg\{\frac{1}{48L_{K}^{2}}, \frac{1}{3\bar{c}_{\eta}}\bigg\}, \nonumber\\
    & \bar{c}_{\eta}=\max \bigg\{32 L_{K}^{2}, 160 L_{K}^{2} m, \frac{24\left(\mu_{g}+L_{g}\right) L_{K}^{2} c_{\beta}}{\bar{c}_{y}}\bigg\}, \nonumber\\
    & \bar{c}_{\gamma}=\max \bigg\{32 L_{g}^{2}, 160 L_{g}^{2} m, \frac{24\left(\mu_{g}+L_{g}\right) L_{g}^{2} c_{\beta}}{\bar{c}_{y}}\bigg\}.
    \end{align*}
    Let $\mathfrak{B}_{t}\triangleq l\left(\bar{\x}_{t}\right)+\bar{c}_{y}\left\|\y_{t}-\y_{t}^{*}\right\|^{2}+\bar{c}_{x}\left\|\x_{t}-\1 \otimes \bar{\x}_{t}\right\|^{2}+\bar{c}_{u}\left\|\u_{t}-\1 \otimes \bar{\u}_{t}\right\|^{2}$ with $\bar{c}_{x}=\frac{6}{(1-\lambda) \alpha_{t}}$, $\bar{\u}_{t}\triangleq \frac{1}{m} \sum_{i=1}^{m} \u_{i,t}$ and $\u_{t}\triangleq  \left[\u_{1,t}^{\top}, \ldots, \u_{m,t}^{\top}\right]^{\top}$.
    If $\alpha_{t}\le \min\{\sqrt{\frac{m}{2 L_{l}^{2}}},
    $ $\frac{\bar{c}_{u}\left( 1-\lambda \right)^{2}}{30},$ $\frac{\bar{c}_{u}(1-\lambda) L_{\mu_{g}} c_{\beta}}{40 L_{y}^{2} \bar{c}_{y}},$ $\frac{\bar{c}_{u}(1-\lambda) \bar{c}_{\eta}}{80 L_{K}^{2}},$ $\frac{\bar{c}_{u}(1-\lambda) \bar{c}_{\gamma}}{80 L_{g}^{2}},$ $\frac{1}{5L_{l}},$ $\frac{1}{3L_{f}},$ $\sqrt{\frac{(1-\lambda)^{2}}{120 L_{K}^{2}}},$ $\frac{1-\lambda}{240 \bar{c}_{u} L_{K}^{2} m},$ $\frac{\bar{c}_{y}(1-\lambda)}{36\left(\mu_{g}+L_{g}\right) \bar{c}_{u} L_{K}^{2} c_{\beta}},$ $1-\lambda \}$ 
    and $\beta_{t}\le \frac{1}{\mu_{g}+L_{g}}$, 
    then the sequence $\left\{ \x_{t},\y_{t} \right\}$ generated by Algorithm~\ref{alg} satisfies:
    \begin{multline*}
    \frac{1}{T}\sum_{t=0}^{T-1}\mathbb{E}\left[ \mathfrak{M}_{t}\right] = \mathcal{O}\left( \frac{\mathfrak{B}_{0}-l^{*}}{T^{2/3}} \right)+\mathcal{O}\left( \frac{ \log\left( T \right)\sigma_{f}^{2} }{T^{2/3}} \right)\\
    +\mathcal{O}\left( \frac{ \log\left( T \right)\sigma_{g}^{2} }{T^{2/3}} \right) = \tilde{\mathcal{O}} \left( \frac{1}{T^{2/3}} \right).
    \end{multline*}
\end{theorem}

Theorem~\ref{thm1} indicates that the decreasing step-sizes $\{\alpha_t, \beta_t\}$ depend on the Lipschitz constants, the number of agents, and the network topology. 
Note also that the choice of step-size $\alpha_t$ is directly affected by $\lambda$, the second largest eigenvalue in magnitude of the weight matrix $\mathrm{\M}$.
Further, Theorem~\ref{thm1} immediately implies the following sample complexity and communication complexity of \algns:
\begin{corollary}[Sample and Communication Complexities of \algns]
    Under the conditions stated in Theorem~\ref{thm1}, \alg requires $\mathcal{O}(\epsilon^{-3/2})$ in  sample complexity and communication complexity to reach an $\epsilon$-stationary point. 
\end{corollary}

\subsection{Proofs of the Main Theoretical Results}

Due to space limitation, we provide a proof sketch of Theorem~\ref{thm1}, which is organized into several key steps:

\smallskip
{\em Step 1) Per-iterate descent of the upper-level objective function:}
We first bound the per-iterate descent of the upper-level objective function as follows:

\begin{lemma}\label{lem_upper_level}
    Under Assumptions~\ref{asmp_up}--\ref{asmp_lo}, the following inequality holds for the consecutive iterates of Algorithm~\ref{alg}:
    \begin{equation*}
        \begin{aligned}
            \mathbb{E}\left[ l\left( \bar{\x}_{t+1} \right)\right.&\left.-l\left( \bar{\x}_{t} \right) \right]\le
             \mathbb{E}\left[ -\frac{\alpha_{t}}{2}\left\|\nabla l\left(\bar{\x}_{t}\right)\right\|^{2}+\frac{2 \alpha_{t}}{m} \sum_{i=1}^{m}\left\|\e_{i, t}^{f}\right\|^{2}\right.\\
            &\left.-\left(\frac{\alpha_{t}}{2}-\frac{L_{l} \alpha_{t}^{2}}{2}\right)\left\|\bar{\u}_{t}\right\|^{2}+\frac{2L_{l}^{2}\alpha_{t}}{m} \sum_{i=1}^{m} \left\|\bar{\x}_{t}-\x_{i, t}\right\|^{2}\right. \\
            &\left. +\frac{2L_{f}^{2} \alpha_{t}}{m} \sum_{i=1}^{m} \left\|\y_{i, t}^{*}-\y_{i, t}\right\|^{2}+\frac{2 \alpha_{t}}{m} \sum_{i=1}^{m}\left\|\b_{i, t}\right\|^{2} \right],
        \end{aligned}
    \end{equation*}
    for all $t\in \left\{ 0,1,\ldots,T-1  \right\}$, where the expectation is taken over all randomness of the algorithm.
\end{lemma}
Lemma~\ref{lem_upper_level}  bounds the expected per-iterate descent of the upper-level objective value, which depends on i) the consensus error of the upper-level parameters $\mathbb{E}[\left\|\bar{\x}_{t}-\x_{i, t}\right\|^{2}]$, 
ii) the momentum-based gradient estimation error of the upper-level objective function $\mathbb{E}[\|\e_{i, t}^{f} \|^{2}]$ including the bias $\|\b_{i, t}\|^{2}$, 
and iii) the approximation gap of the lower-level optimal parameter $\mathbb{E}[\|\y_{i, t}^{*}-\y_{i, t} \|^{2}]$,
which will be bounded in Step 2).

\smallskip
{\em Step 2) Error bound on $\y^{*}(\x)$:}
We show that the approximation error of $\y^*(\x)$ can be bounded as:
\begin{lemma}\label{lem_err_y}
    Under Assumptions~\ref{asmp_up} and \ref{asmp_lo}, the following approximation gap of $\y^*(\x)$ holds for Algorithm~\ref{alg}:
        \begin{align}
    &\mathbb{E}\left[\left\|\y_{i, t+1}-\y_{i, t+1}^{*}\right\|^{2}\right] \le \mathbb{E}\left[\left(1+\frac{1}{c_{1}}\right) L_{y}^{2}\left\| \x_{i,t+1}-\x_{i,t} \right\|^{2}\right.\nonumber\\
            &\left.+\left(1+c_{1}\right)\left(1+c_{0}\right)\left(1- \frac{2 \beta_{t}\mu_{g} L_{g}}{\mu_{g}+L_{g}}\right)\left\|\y_{i, t}-\y_{i, t}^{*}\right\|^{2}\right. \nonumber\\
            &\left.+\left(1+c_{1}\right)\left(1+c_{0}\right)\left(\beta_{t}^{2}- \frac{2 \beta_{t}}{\mu_{g}+L_{g}}\right)\left\|\nabla_{\y} g_{i}\left(\x_{i, t}, \y_{i, t}\right)\right\|^{2}\right. \nonumber\\
            &\left.+\left(1+c_{1}\right)\left(1+\frac{1}{c_{0}}\right) \beta_{t}^{2}\left\|\e_{i, t}^{g}\right\|^{2} \right], \nonumber
        \end{align}
    for all $i\in \{1,\ldots,m\}$ and $t\in \left\{ 0,1,\ldots,T-1  \right\}$ with some positive constants $c_{0},c_{1}> 0$, where the expectation is taken over all randomness of the algorithm.
\end{lemma}
Lemma~\ref{lem_err_y} indicates that the approximation error $\y^*(\x)$ shrinks if $(1+c_{1}) (1+c_{0})(1-2 \beta_{t} \mu_{g} L_{g}/(\mu_{g}+L_{g}))<1$, and is influenced by the momentum-based stochastic gradient estimation error of the lower-level objective function $\mathbb{E}[\|\e_{i, t}^{g} \|^{2}]$. 
Because of the tightly coupled structure of the bilevel problem, $\mathbb{E}[\|\y_{i, t+1}-\y_{i, t+1}^{*} \|^{2}]$ is also affected by the upper-level parameters $\mathbb{E}[ \|\x_{i, t+1}-\x_{i, t} \|^{2}]$, which is in turn affected by the consensus error in decentralized optimization.

\smallskip
{\em Step 3) Shrinking rate of $\e_{i,t}^f$:}
Next, we bound the stochastic gradient estimation error $\e_{i,t}^f$ of the upper-level objective function as follows:

\begin{lemma}\label{lemma:error_f}
    Under Assumptions~\ref{asmp_up}-\ref{asmp_sg}, the stochastic gradient estimation error of the upper-level objective function $\e_{i,t}^f$ satisfies the following relationship:
    \begin{align*}
            \mathbb{E}[\|\e_{i, t+1}^{f}\|^{2}] \leqslant & \mathbb{E} [ (1-\eta_{t+1} )^{2}  \|\e_{i, t}^{f} \|^{2}+2 \eta_{t+1}^{2} \sigma_{f}^{2} \\
            & +4 (1-\eta_{t+1} )^{2} L_{K}^{2}  \|\x_{i, t+1}-\x_{i, t} \|^{2} \\
            & +8 (1-\eta_{t+1})^{2} L_{K}^{2} \beta_{t}^{2} \|\e_{i, t}^{g} \|^{2} \\
            & +8 (1-\eta_{t+1})^{2} L_{K}^{2} \beta_{t}^{2} \|\nabla_{\y} g_{i} (\x_{i,t}, \y_{i,t}) \|^{2} ],
        \end{align*}
    for all $i\in \{ 1,\ldots,m \}$ and $t\in \{ 0,1,\ldots,T-1 \}$, where the expectation is taken over all randomness of the algorithm.
\end{lemma}

Lemma~\ref{lemma:error_f} indicates that the momentum-based stochastic gradient estimation error of the upper-level function $\mathbb{E}[ \| \e_{i,t+1}^{f} \|^{2}]$ is affected by the consensus error of the upper-level parameters, which is contained in $\mathbb{E}[\left\|\x_{i, t+1}-\x_{i, t}\right\|^{2}]$, the lower-level momentum-based stochastic gradient estimation error $\mathbb{E}[ \| \e_{i,t}^{g}  \|^{2}]$, and the full gradient norm $\mathbb{E}[ \|\nabla_{\y} g_{i} (\x_{i,t}, \y_{i,t} ) \|^{2}]$.

\smallskip
{\em Step 4) Shrinking rate of $\e_{i,t}^g$:}
Next, we bound the stochastic gradient estimation error $\e_{i,t}^g$ of the lower-level objective function as follows:

\begin{lemma}\label{lemma:error_g}
    Under Assumptions~\ref{asmp_up}-\ref{asmp_sg}, the stochastic gradient estimation error of the lower-level function satisfies the following relationship:
    \begin{align*}
            \mathbb{E} [ \|\e_{i, t+1}^{g} \|^{2} ] &\le \mathbb{E}[ (1-\gamma_{t+1} )^{2}  (1+8 L_{g}^{2} \beta_{t}^{2} )  \|\e_{i, t}^{g} \|^{2} \\
            & +4 (1-\gamma_{t+1})^{2} L_{g}^{2} \|\x_{i, t+1}-\x_{i, t} \|^{2}+2 r_{t+1}^{2} \sigma_{g}^{2} \\
            & +8 (1-\gamma_{t+1} )^{2} L_{g}^{2} \beta_{t}^{2}  \|\nabla_{\y} g_{i} (\x_{i, t}, \y_{i, t} ) \|^{2} ],
        \end{align*}
    for all $i\in \{1,\ldots,m \}$ and $t\in \{ 0,1,\ldots,T-1 \}$, where the expectation is taken over all randomness of the algorithm.
\end{lemma}

Lemma~\ref{lemma:error_g} shows that, since decentralized bilevel optimization is a composition of a lower-level problem and an upper-level problem, the momentum-based stochastic gradient estimation error of the lower-level function $\mathbb{E}[\| \e_{i,t+1}^{g}  \|^{2}]$ is influenced by the consensus error of the upper-level parameters, which is contained in $\mathbb{E}[ \|\x_{i, t+1}-\x_{i, t} \|^{2}]$.

\smallskip
{\em Step~5) Iterate contractions:}
Next, we establish the following iterate contraction results of the \alg algorithm:

\begin{lemma}\label{lemma:iterates}
    The following contraction properties of the iterates in Algorithm~\ref{alg} hold:
    \begin{align*}
        &\left\|\x_{t+1}-\1 \otimes \bar{\x}_{t+1}\right\|^{2} \le \left(1+c_{2}\right) \lambda^{2}\left\|\x_{t}-\1 \otimes \bar{\x}_{t}\right\|^{2}\\
        &\qquad\qquad\qquad\qquad\qquad+\left(1+\frac{1}{c_{2}}\right) \alpha_{t}^{2}\left\|\u_{t}-\1 \otimes \bar{\u}_{t}\right\|^{2},\\
        &\left\|\u_{t+1}-\1 \otimes \bar{\u}_{t+1}\right\|^{2}\le \left(1+c_{3}\right) \lambda^{2}\left\|\u_{t}-\1 \otimes \bar{\u}_{t}\right\|^{2}\\
        &\qquad\qquad\qquad\qquad\qquad+\left(1+\frac{1}{c_{3}}\right)\left\|\p_{t+1}-\p_{t}\right\|^{2},
    \end{align*}
    where $c_{2},c_{3}> 0$ are constants, and $\p_{t}\triangleq\left[\p_{1, t}^{\top}, \cdots, \p_{m, t}^{\top}\right]^{\top}$. 
    In addition, we have
    \begin{align*}
        &\left\|\x_{t+1}-\x_{t}\right\|^{2}\leqslant 8\left\|\x_{t}-\1 \otimes \bar{\x}_{t}\right\|^{2}\\
        &\qquad\qquad\qquad\quad +4 \alpha_{t}^{2}\left\|\u_{t}-\1 \otimes \bar{\u}_{t}\right\|^{2}+4 \alpha_{t}^{2} m\left\|\bar{\u}_{t}\right\|^{2},\\
        &\mathbb{E} [ \| \p_{t+1}-\p_{t} \|^{2} ] \le \mathbb{E}\left[3 \eta_{t+1}^{2} \sum_{i=1}^{m} \|\e_{i, t}^{f} \|^{2}+24 L_{K}^{2} m \alpha_{t}^{2}  \|\bar{\u}_{t} \|^{2} \right. 
        \\
        &\left. +12 L_{K}^{2} \beta_{t}^{2} \sum_{i=1}^{m}  \|\nabla_{\y} g_{i} (\x_{i, t}, \y_{i,t}) \|^{2}+48 L_{K}^{2} \|\x_{t}-\1 \otimes \bar{\x}_{t} \|^{2} +\right.\\
        &\left. 24 L_{K}^{2} \alpha_{t}^{2} \|\u_{t}\!-\!\1 \otimes \bar{\u}_{t}\|^{2}\!+\!12 L_{K}^{2} \beta_{t}^{2} \sum_{i=1}^{m} \|\e_{i, t}^{g}\|^{2} \!+\! 3 \eta_{t+1}^{2} \sigma_{f}^{2} m\right],
    \end{align*}
    where $\mathbb{E}[\cdot]$ is taken over all randomness of the algorithm.
\end{lemma}

Note that we only attempt to reach consensus in the upper-level $\x$-variables.
Step~5 quantifies the impacts of the consensus error in the iterates of the upper-level $\x$-variables, which is important to analyze the convergence of \algns.
Our key idea is to first define $\tilde{\mathrm{\M}}=\mathrm{\M} \otimes \mathrm{\I}_{m}$.
Since $\x_{t}-\1 \otimes \x_{t}$ is orthogonal to $\1$, which is the eigenvector corresponding to the largest eigenvalue of $\tilde{\mathrm{\M}}$, and $\lambda=\max\left\{ \left| \lambda_{2} \right|,\left| \lambda_{m} \right| \right\}$, we have $\|\tilde{\mathrm{\M}} \x_{t}-\1 \otimes \bar{\x}_{t}\|^{2}= \|\tilde{\mathrm{\M}} (\x_{t}-\1 \otimes \bar{\x}_{t} ) \|^{2} \leq \lambda^{2} \|\x_{t}-\1 \otimes \bar{\x}_{t} \|^{2}$.

\smallskip
{\em Step 6) Decrement of a constructed potential function:}
Next, we define a potential function $W_{t}$ as follows:
\begin{multline} \label{potential_function}
    W_{t} = l(\bar{\x}_{t})+\bar{c}_{y} \|\y_{t}-\y_{t}^{*} \|^{2}+\bar{c}_{x}  \|\x_{t}-\1 \otimes \bar{\x}_{t} \|^{2} \\
    +\bar{c}_{u} \|\u_{t}-\1 \otimes \bar{\u}_{t} \|^{2}+\frac{ \|\e_{t}^{f} \|^{2}}{\bar{c}_{\eta}\alpha_{t-1}}+\frac{ \|\e_{t}^{g} \|^{2}}{\bar{c}_{\gamma}\alpha_{t-1}}.
\end{multline}
Then, we can show the following decrement results for $\{W_t\}$:
\begin{lemma}\label{lemma:descent_potential_function}
    Choose $c_{0}=\frac{\beta_{t} L_{\mu_{g}}}{1-2 \beta_{t} L_{\mu_{g}}}$, $c_{1}=\frac{\beta_{t} L_{\mu_{g}}}{2(1-\beta_{t} L_{\mu_{g}})}$ and $c_{2}=c_{3}=\frac{1}{\lambda}-1$.
    Under the same conditions as in~Theorem~\ref{thm1} and using the results in Lemmas~\ref{lem_upper_level}--\ref{lemma:iterates}, the iterates generated by Algorithm~\ref{alg} satisfy:
    \begin{equation*}
    \begin{aligned}
        \mathbb{E}\left[ W_{t+1}\right. & \left.-W_{t} \right] \leq \mathbb{E}\left[-\frac{\alpha_{t}}{2}\left\|\nabla l\left(\bar{\x}_{t}\right)\right\|^{2}-\frac{2 L_{f}^{2}}{m} \alpha_{t}\left\|\y_{t}^{*}-\y_{t}\right\|^{2}\right.\\
        &\left.-\frac{1}{1-\lambda}\left\|\x_{t}-\1 \otimes \bar{\x}_{t}\right\|^{2}+\frac{2 \alpha_{t}}{m} \sum_{i=1}^{m}\left\|\b_{i,t}\right\|^{2}\right. \\
        &\left.+\frac{2m\eta_{t+1}^{2}}{\bar{c}_{\eta}\alpha_{t}}\sigma_{f}^{2}+\frac{2m\gamma_{t+1}^{2}}{\bar{c}_{\gamma}\alpha_{t}}\sigma_{g}^{2}+\frac{3\bar{c}_{u}m\eta_{t+1}^{2}}{\alpha_{t}}\sigma_{f}^{2}\right].
    \end{aligned}
    \end{equation*}
\end{lemma}
With the proposed potential function and setting the parameters properly, we can make the coefficients of $\mathbb{E}[ \|\u_{t}-\1 \otimes \bar{\u}_{t}  \|^{2}]$, $\mathbb{E}[ \|\bar{\u}_{t} \|^{2}]$, $\mathbb{E}[\sum_{i=1}^{m} \|\e_{,t}^{f} \|^{2}]$, $\mathbb{E}[\sum_{i=1}^{m} \|\e_{,t}^{g} \|^{2}]$ and $\mathbb{E}[\sum_{i=1}^{m} \|\nabla _{\y}g_{i} ( \x_{i,t},\y_{i,t} ) \|^{2}]$ to be non-positive in the range of $\alpha_{t}$ and $\beta_{t}$, which leads to the stated result in Lemma~\ref{lemma:descent_potential_function}.

\smallskip
{\em Step 7) Proof of Theorem~\ref{thm1}:}
Note that $\sum_{t=0}^{T-1}\alpha_{t}^{3}\le \log\left( T+1 \right)$ with $\omega\ge 1$.
Telescoping the result in Lemma~\ref{lemma:descent_potential_function} from $0$ to $T-1$, and multiplying by $2/\alpha_{T}$ on both sides yields: 
\begin{align*}
&\frac{1}{T} \sum_{t=0}^{T-1} \mathbb{E}\left[\mathfrak{M}_{t}\right] \leqslant {\left[\frac{2\left(\mathfrak{B}_{0}-l^{*}\right)}{\alpha_{T} T}+\frac{4}{m \alpha_{T} T} \sum_{t=0}^{T-1} \alpha_{t} E\left[\sum_{i=1}^{m}\left\|\b_{i,t}\right\|^{2}\right]\right.} \\
&+\frac{4 c_{\eta}^{2}m\sigma_{f}^{2}}{\bar{c}_{\eta}} \frac{\log (T+1)}{\alpha_{T} T}+\frac{4 c_{\gamma}^{2}m\sigma_{g}^{2}}{\bar{c}_{\gamma}} \frac{\log (T+1)}{\alpha_{T} T}+\frac{2\sigma_{f}^{2}}{\bar{c}_{\eta}\alpha_{-1}\alpha_{T} T} \\
&\left.+6\bar{c}_{u}c_{\eta}^{2}m\sigma_{f}^{2}\frac{\log (T+1)}{\alpha_{T} T}+\frac{2\sigma_{g}^{2}}{\bar{c}_{\gamma}\alpha_{-1}\alpha_{T} T}\right] / \min \left\{\frac{4 L_{f}^{2}}{m}, \frac{2}{1-\lambda}\right\}.
\end{align*}
Using the fact that $\left\| \b_{i,t} \right\|=1/T$ when  $K=\left( \mu_{g}/L_{g} \right)\log\left( C_{g_{xy}}C_{f_{y}}T/\mu_{g} \right)$, we have:
\begin{equation*}
    \frac{1}{T} \sum_{t=0}^{T-1} \mathbb{E}\left[\mathfrak{M}_{t}\right] = \mathcal{O}\left(\frac{\mathfrak{B}_{0}-l^{*}}{T^{2 / 3}}\right)+\widetilde{\mathcal{O}}\left(\frac{\sigma_{f}^{2}}{T^{2 / 3}}\right)+\widetilde{\mathcal{O}}\left(\frac{\sigma_{9}^{2}}{T^{2 / 3}}\right),
\end{equation*}
where $\mathfrak{B}_{t}\triangleq l (\bar{\x}_{t} ) + \bar{c}_{y} \|\y_{t}-\y_{t}^{*} \|^{2}+\bar{c}_{x} \|\x_{t}-\1 \otimes \bar{\x}_{t} \|^{2}+\bar{c}_{u} \|\u_{t}-\1 \otimes \bar{\u}_{t} \|^{2}$.
This completes the proof of Theorem~\ref{thm1}.


\section{Numerical Results} \label{sec: NumericalResults}

\begin{figure*}[htbp]
	\centering
	\begin{minipage}[t]{0.48\textwidth}
		\begin{subfigure}[t]{0.48\textwidth}
        \centering
        \includegraphics[width=1\linewidth]{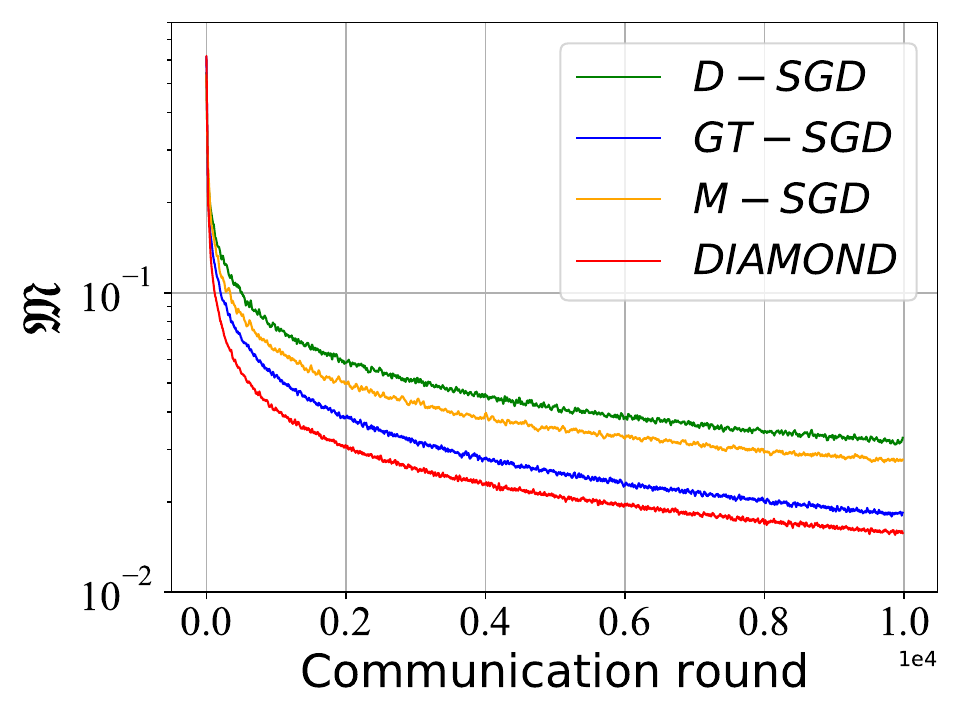}
        \caption{MNIST dataset.}
    \end{subfigure}
    \hspace{1mm}
    \begin{subfigure}[t]{0.48\textwidth}
        \centering
        \includegraphics[width=1\linewidth]{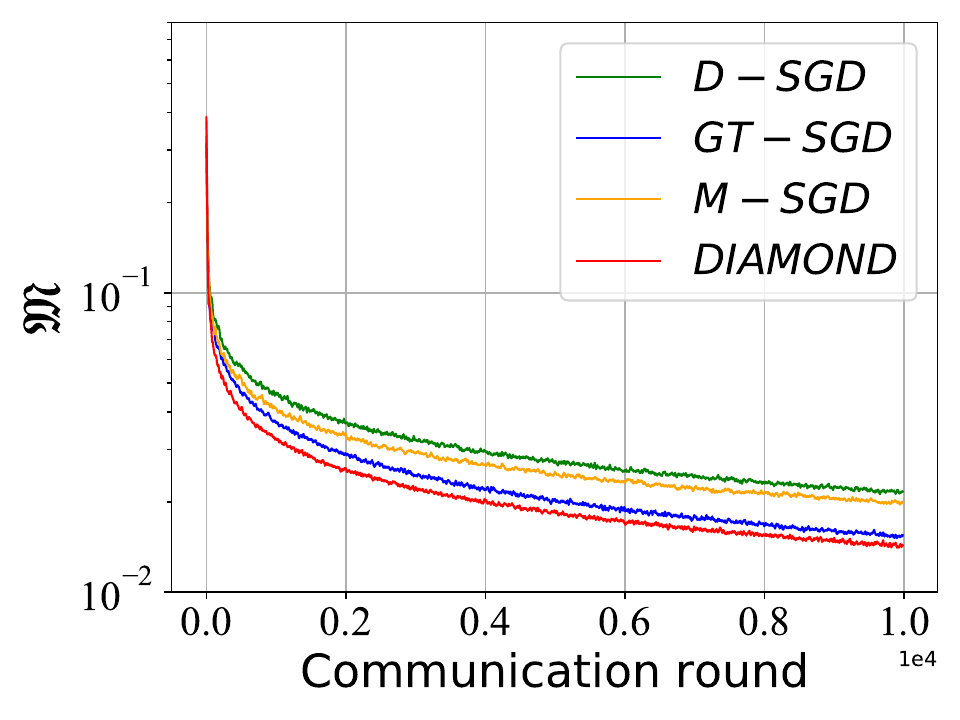}
        \caption{CIFAR-10 dataset.}
    \end{subfigure}
    \caption{Convergence performance on the meta-learning problem with 9-agent network.}
    \label{fig:convergence_compar_9}
	\end{minipage}
	\hspace{.01\textwidth}
	\begin{minipage}[t]{0.48\textwidth}
		\begin{subfigure}[t]{0.48\textwidth}
        \centering
        \includegraphics[width=1\linewidth]{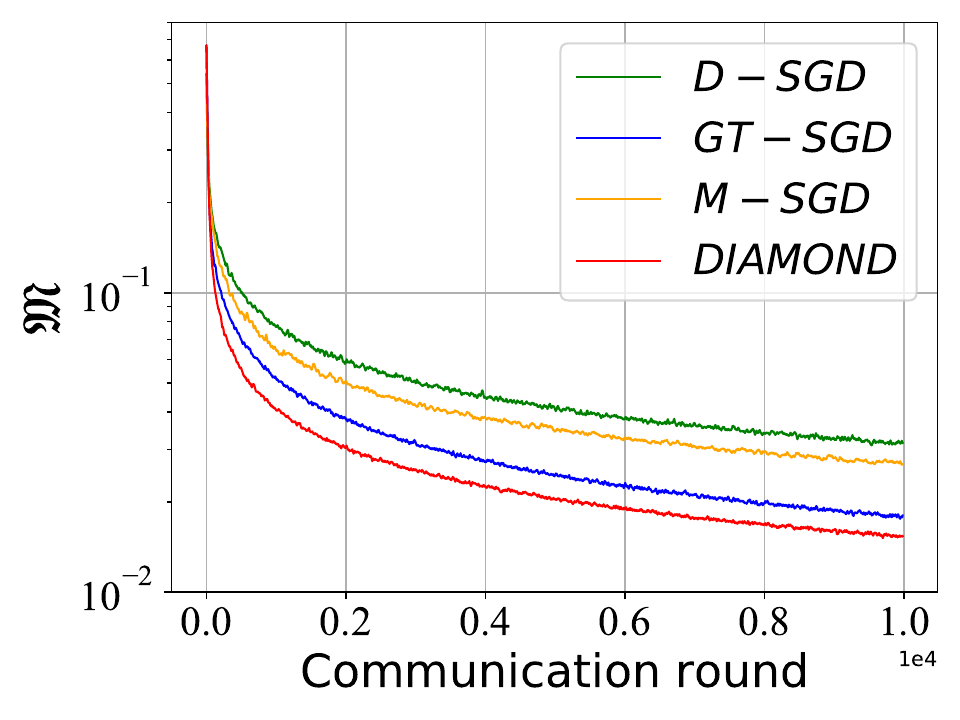}
        \caption{MNIST dataset.}
    \end{subfigure}
    \hspace{1mm}
    \begin{subfigure}[t]{0.48\textwidth}
        \centering
        \includegraphics[width=1\linewidth]{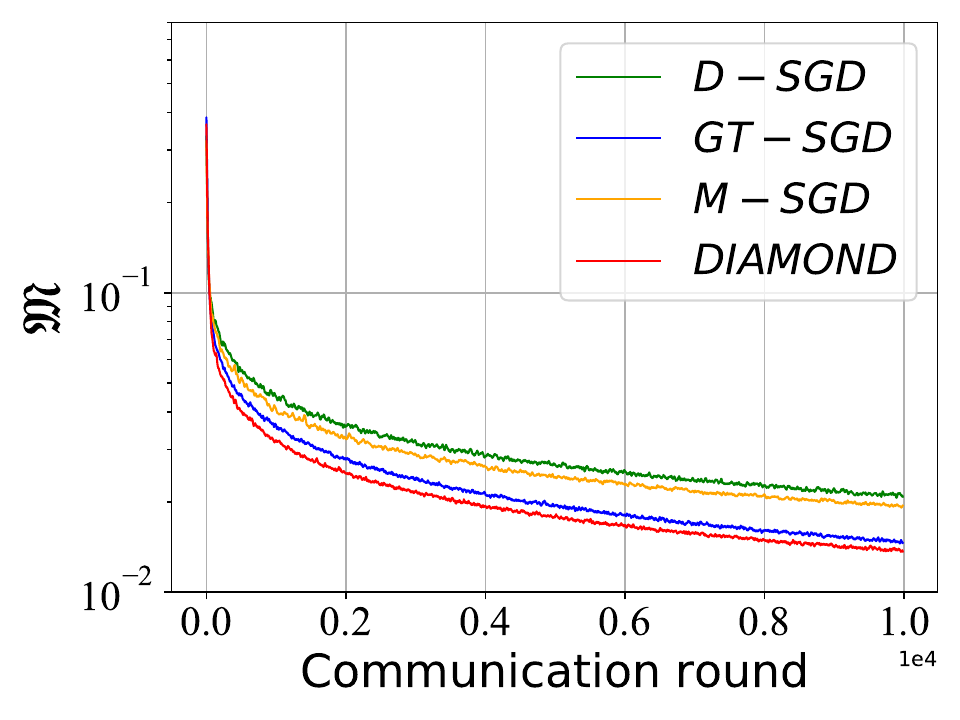}
        \caption{CIFAR-10 dataset.}
    \end{subfigure}
		\caption{Convergence performance on the meta-learning problem with 15-agent network.}
    \label{fig:convergence_compar_15}
	\end{minipage}
	\vspace{-0.1in}
\end{figure*}

\begin{figure*}[htbp]
	\centering
	\begin{minipage}[t]{0.48\textwidth}
		\begin{subfigure}[t]{0.48\textwidth}
        \centering
        \includegraphics[width=1\linewidth]{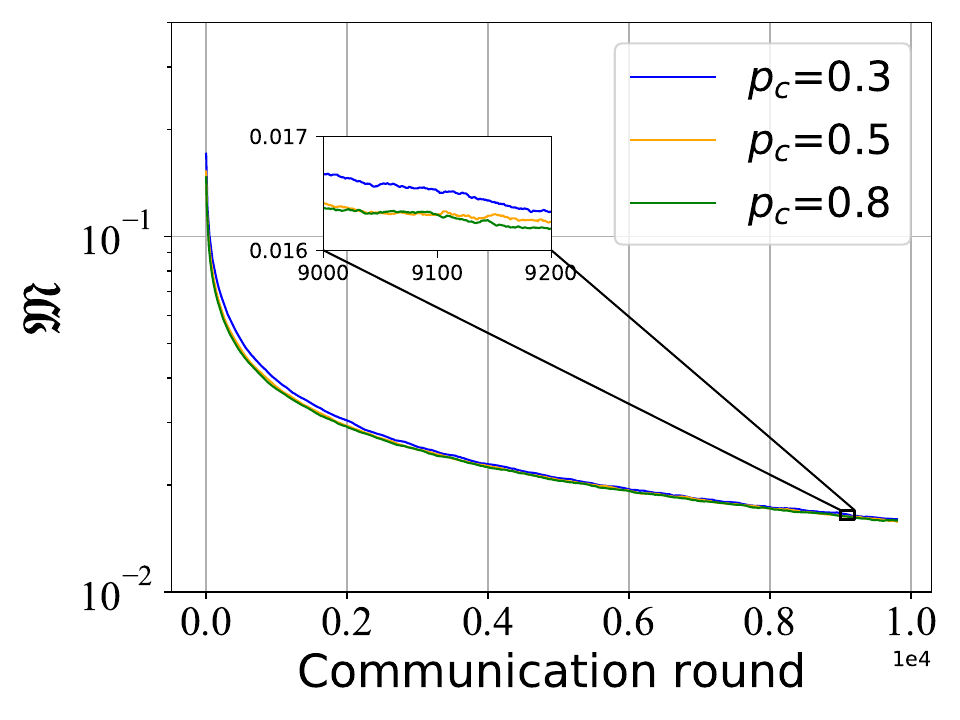}
        \caption{9-agent network.}
    \end{subfigure}
    \hspace{1mm}
    \begin{subfigure}[t]{0.48\textwidth}
        \centering
        \includegraphics[width=1\linewidth]{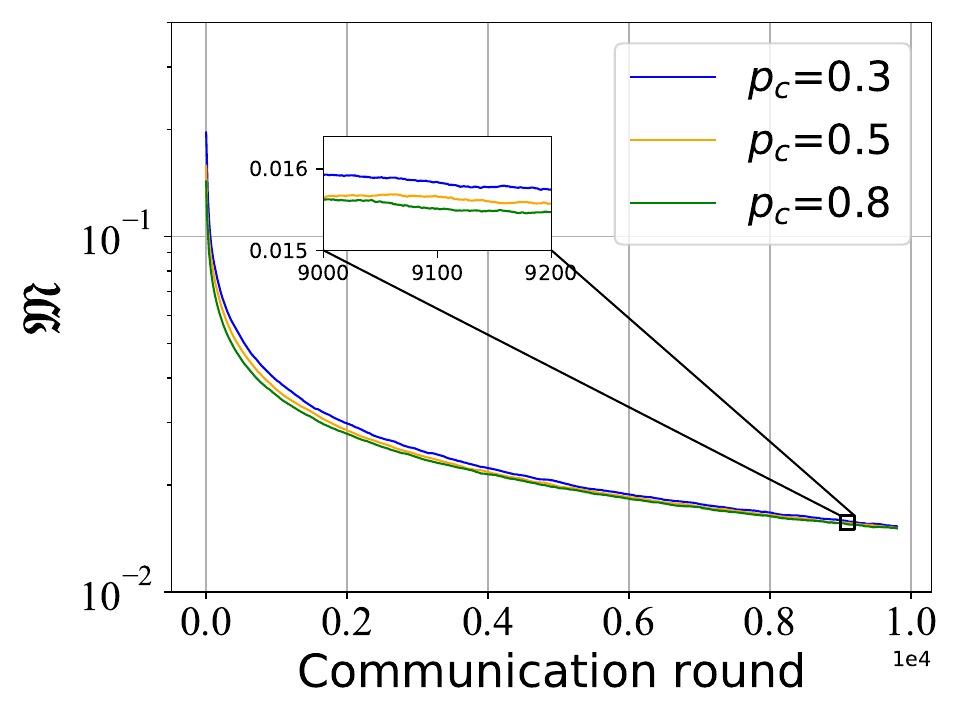}
        \caption{15-agent network.}
    \end{subfigure}
    \caption{Convergence performance with different edge connectivity probabilities $p_c$ on the meta-learning problem.}
    \label{fig:convergence_pc}
	\end{minipage}
	\hspace{.01\textwidth}
	\begin{minipage}[t]{0.48\textwidth}
		\begin{subfigure}[t]{0.48\textwidth}
        \centering
        \includegraphics[width=1\linewidth]{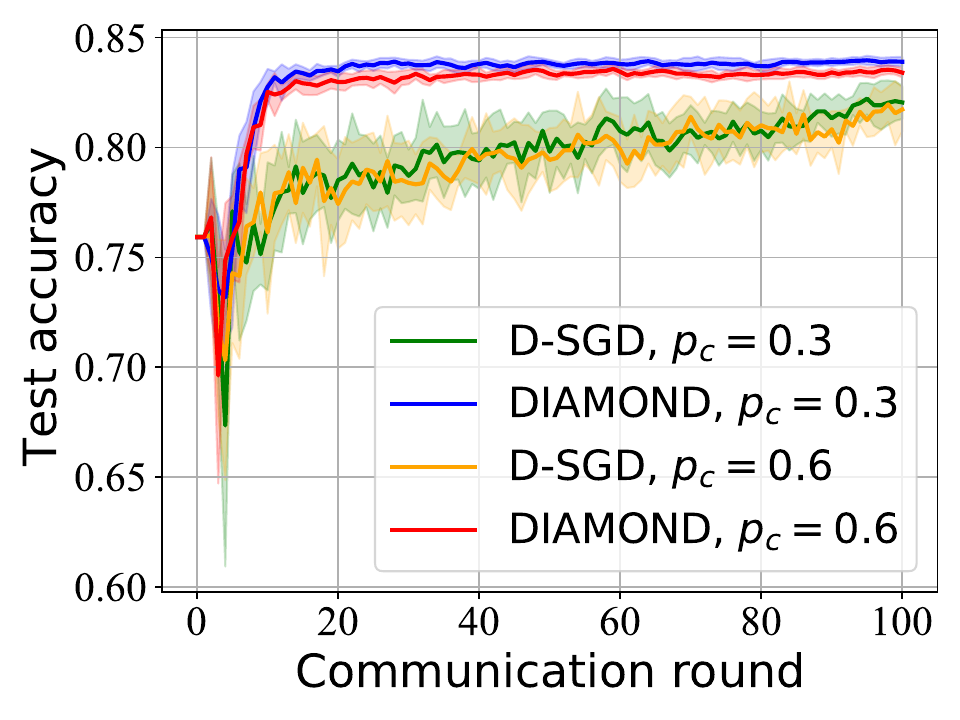}
    \end{subfigure}
    \hspace{1mm}
    \begin{subfigure}[t]{0.48\textwidth}
        \centering
        \includegraphics[width=1\linewidth]{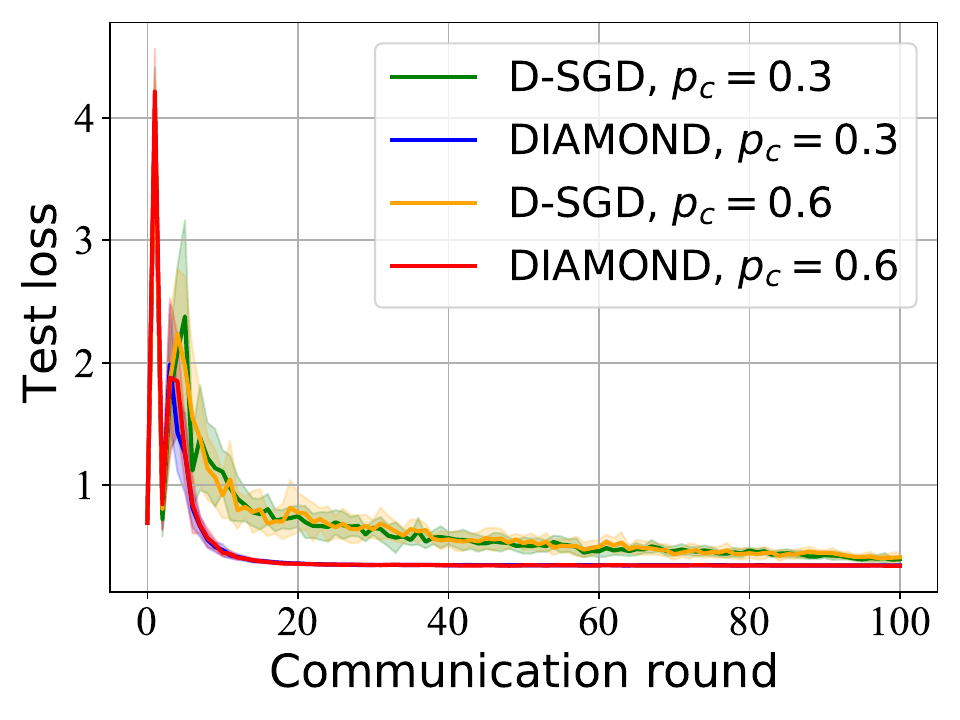}
    \end{subfigure}
		\caption{Comparison between D-SGD and \alg on the hyper-parameter optimization problem.}
    \label{fig:convergence_hyperpara}
	\end{minipage}
	\vspace{-0.2in}
\end{figure*}

In this section, we conduct numerical experiments to verify the theoretical results of the proposed \alg algorithm using decentralized meta-learning problems and hyper-parameter optimization problems.

\smallskip
\textbf{1) Decentralized Meta-Learning:}
The formulation of the decentralized meta-learning problem is the same as \eqref{eq:dec_bilevel_prob}, where $\x$ is the common parameters across all agents and $\y_i$ is the task-specific parameters of $\mathcal{T}_i$. 
The upper-level optimization objective function $f_i(\x,\y_i^*(\x))$ is the loss function related to task $\mathcal{T}_i$ and is non-convex in $\x$. 
The lower-level optimization objective function $g_i(\x,\y_i)$ satisfies the strongly convex requirements in $\y_i$ by using a strongly convex regularizor $\mathcal{R}(\y_i)=\frac{1}{2}\|\y_i\|_2^2$. 
The decentralized meta-learning problem aims to learn the common parameters that can be adapted to specific tasks.
The classifier is based on a two-hidden-layer fully connected neural network for the decentralized meta-learning problem. 
The network topology is generated based on Erdös-Rényi random graph by NetworkX \cite{hagberg2008exploring}. 
We choose the consensus matrix as $\M = \I - \frac{2 \V}{3 \rho_{\max}(\V)}$, where $\V$ is the Laplacian matrix and $\rho_{\max}$ is the largest eigenvalue of $\V$.

\smallskip
{\bf 1-a) Comparison between Decentralized Stochastic Algorithms:}
We conduct decentralized meta-learning on MNIST ~\cite{lecun1998gradient} and CIFAR-10~\cite{krizhevsky2009learning} datasets with 9- and 15-agent networks.
The edge connectivity probability is $p_c=0.3$. 
We compare the proposed \alg algorithm with three decentralized stochastic bilevel optimization methods:
\begin{itemize}
    \item {\bf Decentralized Stochastic Gradient Descent (D-SGD)\cite{nedic2009distributed}:} Each agent directly updates its local copy of the upper-level parameters $\x_{i,t}$ with the stochastic gradient of the upper-level objective function, i.e., $\x_{i,t}=\sum_{j \in \mathcal{N}_{i}}[\M]_{i j} \x_{j, t-1}-\alpha_t\hat{\nabla} f_{i}\left(\x_{i, t}, \y_{i, t} ; \bar{\xi}_{i,t}\right)$ and $\y_{i,t}=\y_{i, t-1}-\beta_t\nabla_{\y} g_{i}\left(\x_{i, t}, \y_{i, t} ; \zeta_{i,t}\right)$.
    
    \item {\bf Gradient-Tracking Stochastic Gradient Descent (GT-SGD):} In addition to the DSGD update, GT-SGD performs gradient tracking over $\x_{i,t}$-variables. 
    Specifically, GT-SGD updates the upper-level variables as: $\x_{i, t+1} \!=\! \sum_{j \in \mathcal{N}_{i}}[\M]_{i j} \x_{j, t} \!-\! \alpha_{t} \u_{i, t}$, where $\u_{i, t} \!=\! \sum_{j \in \mathcal{N}_{i}}[\M]_{i j} \u_{j, t-1} \!+\! \hat{\nabla} f_{i} (\x_{i, t}, \y_{i, t} ; \bar{\xi}_{i,t} ) \!-\! \hat{\nabla} f_{i} (\x_{i, t-1}, \y_{i, t-1} ; \bar{\xi}_{i,t-1} )$. 
    
    \item {\bf Momentum Stochastic Gradient Descent (M-SGD):} This algorithm can be viewed as a simplified version of \alg by neglecting gradient tracking. 
    Specifically, we replace the update step \eqref{eq:x_update} by $\x_{i, t+1}=\sum_{j \in \mathcal{N}_{i}}[\mathrm{\M}]_{i j} \x_{j, t}-\alpha_{t} \p_{i,t}$.
\end{itemize}
We set the learning rates $\alpha_t\!=\!c_{\alpha}(\omega\!+\!t)^{-1/3}$ and $\beta_t\!=\!c_{\beta}\alpha_t$ following Theorem~\ref{thm1}, where $c_{\alpha}\!=\!10$, $\omega\!=\!2$ and $c_{\beta}\!=\!10$. 
The momentum coefficients are chosen as $\eta_{t+1}\!=\!c_{\eta}\alpha_t^2$ and $\gamma_{t+1}\!=\!c_{\gamma}\alpha_t^2$ for M-SGD and \algns, respectively, where $c_{\eta}\!=\!c_{\gamma}\!=\!0.1$. 
Figs.~\ref{fig:convergence_compar_9} and \ref{fig:convergence_compar_15} illustrate that the \alg algorithm outperforms other algorithms for solving decentralized bilevel optimization problems in terms of the convergence metric in both 9-agent and 15-agent networks, which implies lower sample and communication complexity.


\smallskip
{\bf 1-b) Impact of connectivity probability:}
We evaluate the impact of edge connection probability $p_c$ on the performance of \alg with the nine-agent network. 
We choose $p_c$ from $\{0.3,0.5,0.8\}$, and the parameters of learning rate and momentum coefficient are the same as those in the previous setting. 
As shown in Fig.~\ref{fig:convergence_pc}, there is only a slight increase in convergence rate with a higher $p_c$-value, which reflects that the performance of \alg is not sensitive to the edge connection probability, and the proposed \alg algorithm can adapt to various network settings.

\smallskip
{\bf 2) Hyper-parameter optimization.}
Next, we compare \alg with D-SGD using the logistic regression problem~\cite{grazzi2020iteration,ji2021bilevel} with the same formulation as in~\eqref{eq:dec_bilevel_prob},
where 
$f_i(\x,\y_i^*(\x))\!=\!\frac{1}{ |\mathcal{D}_{\text {val},i} |} \sum_{ (\a_{j}, \c_{j} ) \in \mathcal{D}_{\text {val},i}} Q(\a_{j}^T \y_i^{*}, \c_{j} )$,
$g_i(\x,\y_i)\! =\!\frac{1}{ |\mathcal{D}_{\text {tr},i} |} \sum_{ (\a_{j}, \c_{j} ) \in \mathcal{D}_{\text{tr},i}} Q(\a_{j}^T \y_i, \c_{j} )\!+\!\frac{1}{q p} \sum_{k=1}^{q}$  $\sum_{r=1}^{p} \exp(\x_{r} ) \y_{irk}^{2}$. 
Here, $\mathcal{D}_{\text{tr},i}$ and $\mathcal{D}_{\text{val},i}$ are the training and validation datasets for agent $i$, respectively, 
$Q$ is the cross-entropy loss, $q$ is the number of classes, and $p$ is the number of features. 
We use the a9a dataset~\cite{a9a}, where $q\!=\!2$ and $p\!=\!123$, and divide the a9a dataset into training, validation, and testing sets, which contain 40\%, 40\%, and 20\% samples, respectively. 

We compare the proposed \alg algorithm with D-SGD in terms of test accuracy and test error, using five-agent communication networks. 
For both \alg and D-SGD, the learning rates are set as $\alpha_t\!=\!c_{\alpha}{(\omega\!+\!t)}^{-1/3}$ and $\beta_t\!=\!c_{\beta}\alpha_t$, where $c_{\alpha}\!=\!5$, $\omega\!=\!2$, and $c_{\beta}\!=\!1.5$. 
Moreover, the momentum related parameters in \alg are chosen as $\eta_{t+1}\!=\!c_{\eta}\alpha_t^2$ and $\gamma_{t+1}\!=\!c_{\gamma}\alpha_t^2$, where $c_{\eta}\!=\!c_{\gamma}\!=\!0.1$.
As shown in Fig.~\ref{fig:convergence_hyperpara}, \alg has a faster convergence rate than that of D-SGD over various networks with different edge connection probabilities.
This validates the superiority of the proposed \alg algorithm.




\section{Conclusion} \label{sec: Conclusion}
In this paper, we proposed the \alg algorithm for decentralized bilevel optimization with non-convex upper-level subproblems and strongly-convex lower-level subproblems. 
\alg utilizes a network-consensus approach and adopts a single-loop algorithmic architecture along with momentum-based stochastic gradient estimations and gradient tracking techniques.
This is contrast to existing related works that use a double-loop architecture, full gradient estimations, or large-batch gradients.
We showed that \alg achieves $\mathcal{O}\left( \epsilon^{-3/2} \right)$ in both sample and communication complexities to reach an $\epsilon$-stationary point, outperforming  existing works.
We also conducted numerical experiments using meta-learning and  hyper-parameter optimization problems to verify our theoretical findings. 
Future directions include i) to consider Hessian approximation to further reduce computation costs of the Hessian matrix and ii) to extend the proposed algorithm to wireless networks with channel noise and fading. 

\bibliographystyle{IEEEtran}
\bibliography{BIB/BilevelOptimization}


\end{document}